\documentclass{article}

\usepackage{microtype}
\usepackage{graphicx}
\usepackage{booktabs} %
\usepackage{listings}
\usepackage{hyperref}
\usepackage{fvextra}
\usepackage{rotating}

\usepackage[accepted]{icml2025}

\usepackage{amsmath}
\usepackage{amssymb}
\usepackage{mathtools}
\usepackage{amsthm}

\usepackage[capitalize,noabbrev]{cleveref}

\theoremstyle{plain}

\theoremstyle{definition}

\theoremstyle{remark}

\usepackage[textsize=tiny]{todonotes}

\usepackage[utf8]{inputenc} %
\usepackage{hyperref}       %
\usepackage{url}            %
\usepackage{booktabs}       %
\usepackage{amsfonts}       %
\usepackage{pifont}
\usepackage{amssymb}
\usepackage{nicefrac}       %
\usepackage{microtype}      %
\usepackage{xcolor}         %
\usepackage{graphicx}
\usepackage{multirow}
\usepackage{comment}
\usepackage{mathtools}
\usepackage{graphicx}
\usepackage{caption}
\usepackage{subcaption}
\usepackage{tablefootnote}
\usepackage{textcomp} 
\usepackage{scalerel}

\newcommand*{\escape}[1]{\texttt{\textbackslash#1}}

\usepackage{caption}
\usepackage{makecell}

\usepackage{hyperref}
\usepackage{url}
\usepackage{graphicx}
\usepackage{multirow}
\usepackage{enumitem}
\usepackage{cleveref}
\usepackage{xcolor,colortbl}
\usepackage{wrapfig,lipsum,booktabs}
\usepackage{bbm}
\usepackage[most]{tcolorbox}
\usepackage{tikz}
\usetikzlibrary{arrows.meta} %
\usepackage{fontawesome5}

\hypersetup{colorlinks, linkcolor={red!55!black}, citecolor={blue!65!black}, urlcolor={blue!65!black}}

\definecolor{green1}{HTML}{7C9895}
\definecolor{green2}{HTML}{9FBA95}
\definecolor{green3}{HTML}{4F845C}
\definecolor{green4}{HTML}{B2D3A4}
\definecolor{green5}{HTML}{C9DCC4}

\definecolor{jx}{rgb}{0.9, 0, 0.1}

\newcommand{\method}{\textsl{RIP}}
\newcommand{\methodlong}{Rejecting Instruction Preferences}
\newcommand{\para}[1]{\textbf{#1~~}}

\icmltitlerunning{\methodlong{}  (\method)}

\usepackage{colortbl}
\usepackage{xcolor}

\newcommand{\colorIntensity}[3]{%
    \ifdim #1 pt > #2 pt
        \cellcolor{green!\numexpr(#1-#2)/(#3-#2)*100\relax}#1
    \else
        \cellcolor{red!\numexpr(#2-#1)/(#2-#3)*100\relax}#1
    \fi
}

\begin{document}

\twocolumn[

\icmltitle{R.I.P.~\faSkull: Better Models by Survival of the Fittest Prompts}

\author{%
   Ping Yu $^{1}$ \quad Weizhe Yuan$^{1,2}$  \quad {Olga Golovneva}$^{1}$  \quad {Tianhao Wu}$^{3}$   
   \\ \quad {\bf Sainbayar Sukhbaatar}$^{1}$  \quad     \textbf{Jason Weston Xu}$^{1}$ 
   \quad \textbf{Jing Xu}$^{1,2}$   \\\\
   $^1$ Meta ~~~~~~~~~~ $^{2}$ NYU  ~~~~~~~~~ $^{2}$ UC Berkeley \\
 }
\begin{icmlauthorlist}
\icmlauthor{Ping Yu}{meta}
\icmlauthor{Weizhe Yuan}{meta,nyu}
\icmlauthor{Olga Golovneva}{meta}
\icmlauthor{Tianhao Wu}{ucb} \\
\icmlauthor{Sainbayar Sukhbaatar}{meta}
\icmlauthor{Jason Weston}{meta,nyu}
\icmlauthor{Jing Xu}{meta}
\end{icmlauthorlist}
\icmlaffiliation{meta}{Meta}
\icmlaffiliation{nyu}{New York University}
\icmlaffiliation{ucb}{UC Berkeley}
\icmlcorrespondingauthor{Jing Xu}{jingxu23@meta.com}

\icmlkeywords{Machine Learning, ICML}

\vskip 0.3in
]

\printAffiliationsAndNotice{}

\begin{abstract}
Training data quality is one of the most important drivers of final model quality. In this work, we introduce a method for evaluating data integrity based on the assumption that low-quality input prompts result in high variance and low quality responses. This is achieved by measuring the {\em rejected response quality} and the {\em reward gap} between the chosen and rejected preference pair. Our method, Rejecting Instruction Preferences (\method{}) can be used to filter prompts from existing training sets, or to make high quality synthetic datasets, yielding large performance gains across various benchmarks compared to unfiltered data.
Using Llama 3.1-8B-Instruct, \method{} improves AlpacaEval2 LC Win Rate by 9.4\%, 
Arena-Hard by 8.7\%, and WildBench by 9.9\%.
Using Llama 3.3-70B-Instruct, \method{} improves Arena-Hard from 67.5 to 82.9, which is from 18th place to 6th overall in the leaderboard. Our datasets are available at \href{https://huggingface.co/datasets/facebook/Wildchat-RIP-Filtered-by-8b-Llama}{Wildchat-RIP-Filtered-by-8b-Llama} and \href{https://huggingface.co/datasets/facebook/Wildchat-RIP-Filtered-by-70b-Llama}{Wildchat-RIP-Filtered-by-70b-Llama}.
\end{abstract}

\section{Introduction} \label{sec:intro}

In large language model (LLM) development, a primary driver for advancing
frontier models is curating high-quality training examples. This curation is crucial during both the pretraining \citep{rae2021scaling,touvron2023llama1} and post-training (finetuning) phases \citep{touvron2023llama2}.
Despite the widespread adoption of the ``scaling hypothesis'' \citep{kaplan2020scaling}, 
merely increasing the size of training datasets does not guarantee improved performance if the data are of low quality \citep{chen2023alpagasus,li2023self,zhou2024lima}.
Without sufficient data quality, model training tends not to be fully robust to the associated noise, and final response quality from the model suffers.

\begin{figure}[t!]
    \centering
    \includegraphics[width=0.48\textwidth,trim={0 0 0 0}]{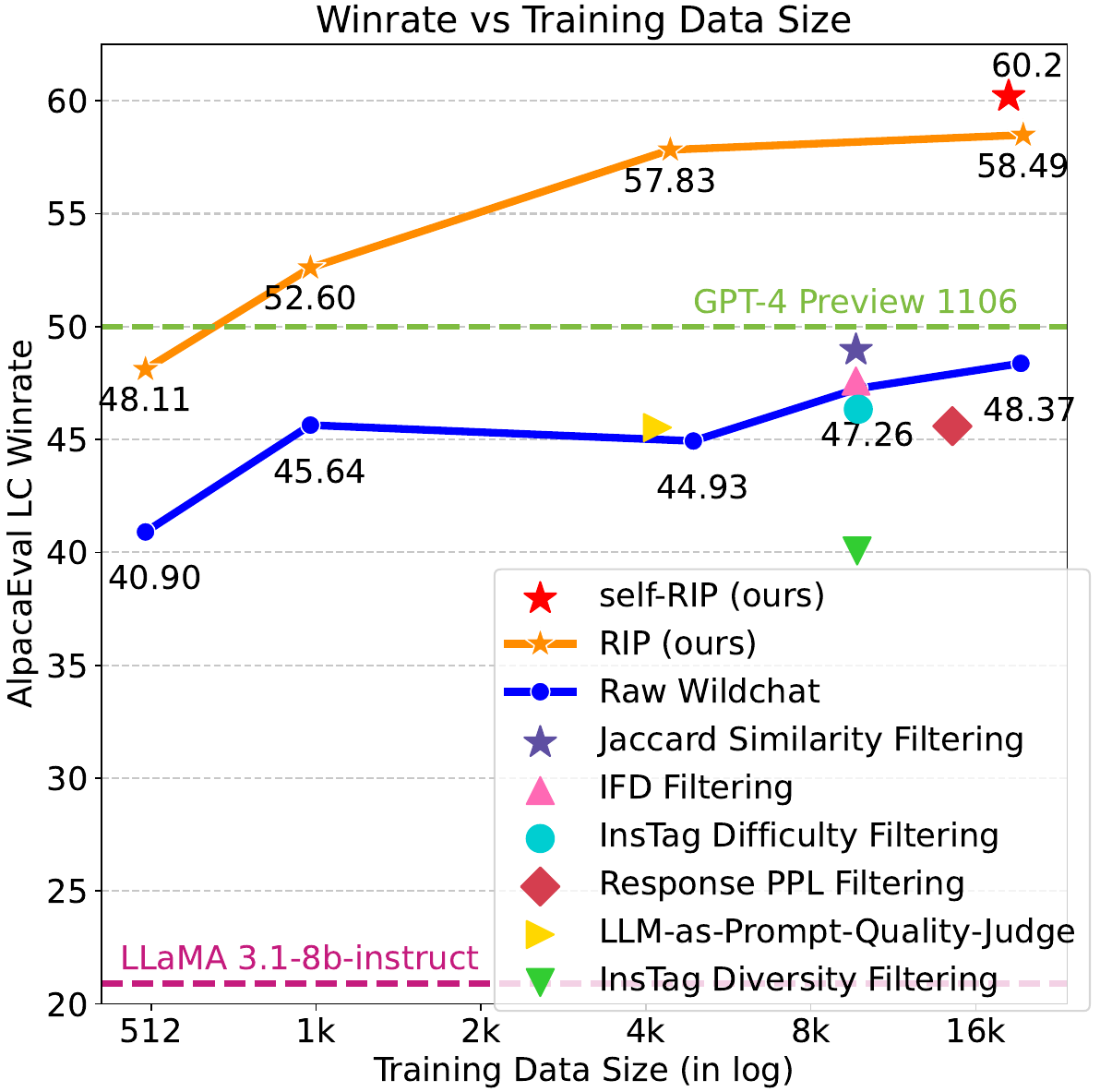}
   \caption{
   { {\bf Our method {\em Rejecting Instruction Preferences} (\method{}) for curating data,  and Self-\method{} for creating synthetic data.}
   The x-axis represents the effective training set size (after filtering). 
   At every data size training on unfiltered WildChat prompts is significantly outperformed  by  \method.
   \method~also outperforms  various other curation baselines.
   Synthetic data built by Self-\method{} improves results further. 
   }} 
   \vspace{-1em}
   \label{fig:data_scaling}
\end{figure}

\begin{table*}[t!]
 \small
\setlength{\tabcolsep}{11pt}
  \centering
\caption{{\bf Rejecting Instruction Preferences (\method{}) and Self-\method{} compared to SOTA models on AlpacaEval2, Arena-Hard and WildBench}. 
By training Llama 3.1-8B-Instruct and Llama 3.3-70B-Instruct on Wildchat instructions curated by \method{}, or synthetic data created by Self-\method{}, our method surpasses many existing SOTA models. 
}
\begin{tabular}{lcccc}
\toprule
 \multicolumn{1}{c}{} &\multicolumn{2}{c}{AlpacaEval2} & \multicolumn{1}{c}{Arena-Hard} & \multicolumn{1}{c}{WildBench}  \\
\cmidrule(lr){2-3} \cmidrule(lr){4-4}  \cmidrule(lr){5-5}
\emph{Standard models}  & \multicolumn{1}{c}{LC Win} & \multicolumn{1}{c}{Win} & \multicolumn{1}{c}{Score}   & Score \\
\midrule
GPT-4 Omni (05/13)   & 57.5 & 51.3 & 74.9 & \textbf{59.3} \\
GPT-4 Turbo (04/09)   & 55.0 & 46.1   & 82.6   & 55.2 \\
Llama 3.1-8B-Instruct & 20.9& 21.8  & 21.3  &  33.1 \\
Llama 3.3-70B-Instruct & 38.9 & 41.5 & 67.5 & 52.8 \\
\midrule
Llama 3.1-8B-Instruct + \method{} (ours) &  57.8 & 57.2   & 43.1  & 45.6\\
Llama 3.1-8B-Instruct + Self-\method{} (ours) & 60.2 & 61.1  &   42.1  &  42.5 \\
Llama 3.3-70B-Instruct + \method{} (ours) & \textbf{67.7} & \textbf{73.2} & \textbf{82.9} & 58.8 \\
\bottomrule
\end{tabular}  \label{tab:alpaca_arena_wildbench_wildchat_sota}
  \vspace{-1em}
\end{table*}

Currently, there are a number of investigated techniques to curate data %
-- most of which are based on heuristics or model judgments given the training inputs. 
In this work, we hypothesize that better judgments of data quality can be made {\em by taking into account the model responses on those data}. 
Specifically, if the prompt is of low quality, then responses exhibit high variability and low quality as well. 
This insight leads us to develop a method for either selecting prompts, or for creating high quality synthetic prompts, both of which yield significant performance gains during post-training.

Our method, \methodlong{} (\method{}), considers the case of instruction finetuning via preference optimization.
It starts with a set of preference pairs consisting of input prompts and chosen and rejected responses.
\method{} considers specific characteristics of the preference pairs, in particular {\em rejected response quality} and the {\em reward gap} between the chosen and rejected preference pair. If the rejected quality is low or the reward gap is high this is an indicator that the prompt is of low quality.

\if 
We thus filter the prompts based on these metrics reward score of the rejected response, its length, and the reward gap between the chosen and rejected responses. The remaining prompts can subsequently be used to fine-tune the model using a preference optimization method like Direct Preference Optimization (DPO) \citep{rafailov2023direct}.
\fi 
We thus filter the prompts based on these metrics. The remaining prompts can subsequently be used to fine-tune the model using RLHF methods like  Direct Preference Optimization (DPO) \citep{rafailov2023direct}, or for creating new synthetic prompts via few-shot prompting. \autoref{tab:alpaca_arena_wildbench_wildchat_sota} illustrates that when trained on Wildchat prompts \citep{zhao2024wildchat} and filtered by \method{}, both Llama 3.1-8B-Instruct and Llama 3.3-70B-Instruct \citep{dubey2024llama} achieve large performance gains, surpassing many state-of-the-art models.

Additionally, we conducted comprehensive experiments comparing the scaling behavior of our data under \method{} filtering with that of unfiltered WildChat raw data, and six alternative filtering methods in \autoref{fig:data_scaling}. 
Our results  demonstrate that \method{} significantly enhances model performance, while other filtering methods yield only marginal improvements. In addition to improvements observed with filtering  human-written data such as Wildchat prompts or HelpSteer2  using different reward signals such as human, classifier or LLM-as-a-Judge, we also show \method{} improves model performance as a method to create synthetic data.

Analysis of our method using t-SNE shows that  \method{}  can eliminate certain undesirable clusters. Additionally, analysis with GPT-4 reveals that \method{} effectively removes  noisy or low quality prompts, ambiguous prompts, unsafe prompts, and examples where preference choices are incorrect.

\section{Related Work}
\para{Data Selection in Pretraining Data}
Given the high variance in quality of pretraining data, data filtering is a critical component for determining  pretrained model quality \cite{hoffmann2022training}. In addition to heuristic preprocessing such as deduplication of similar documents, removal of datasets with heavy test-set overlap, and text extractions from raw Internet content, GPT-3 \citep{brown2020language} applied text filtering to the CommonCrawl dataset based on  similarity to high-quality reference data, significantly reducing final pretraining text data from 45TB down to a 570GB high-quality subset. As language models become more powerful, data curation can also be facilitated by using LLMs as a quality judge. Llama2 and Llama3 employ model-based quality classifiers to filter out non-English and low-quality content from pretraining data \citep{touvron2023llama2,dubey2024llama}. \citet{rae2021scaling,soldaini2024dolma} also demonstrate that applying simple filtering on massive texts brings substantial improvements on downstream performance across the board.

\para{Data Selection in Supervised Fine-Tuning} %
Similarly, post-training also relies on  high-quality data to enhance models' instruction-following capabilities.  Previously, instruction-tuning was regarded as largely dependent on the size of available instruction-tuning examples \cite{mishra2021cross,wei2021finetuned,wang2022super}. More recent work has revealed that training on a smaller yet higher-quality curated set of prompts tends to be more effective in improving models' instruction-following capabilities \citep{zhou2024lima,chen2023alpagasus}. To facilitate data selection, some employ traditional optimization-based data-pruning methods by measuring their impact on model's generalization capabilities \citep{toneva2018empirical,yang2022dataset,xia2024less}. 
Another stream of work studies employing powerful language models to measure the complexity, diversity and quality of instructions \citep{lu2023instag,chen2023alpagasus,touvron2023llama2,dubey2024llama,li2023self}. Alternative filtering approaches proposed automatic metrics such as IFD score \citep{li2023quantity}, or {\small\sc INSTRUCTMINING} which fits a linearly weighted score over a bag of natural language indicators \citep{cao2023instruction} to select examples.

\para{Data Selection in RLHF and Preference Optimization}
The success of preference-optimization methods \citep{stiennon2020learning,rafailov2024direct} has attracted more attention to collecting large scale and high quality preference data. While extensive work shows scaling up preference data through bootstrapping \citep{xu2023some,yuan2024self}, synthesis approaches \citep{lambert2024t,wang2024self}, or crowdsourcing \citep{touvron2023llama2,dubey2024llama}, can boost model performance, the characterization and selection of high-quality pairwise examples is surprisingly underexplored. Most work involving preference optimization  employs  existing methods derived from pretraining and instruction-tuning \citep{touvron2023llama2,dubey2024llama}, such as deduplication, quality classifiers or filtering heuristics. However, such methods overlook the importance of the preference pairs (the chosen and rejected responses). Recent work \citet{wu2024beta,khaki2024rs} shows that preference optimization can be highly sensitive to the choice of response pairs of different reward gaps, focusing more on pair construction than data selection.

\section{Rejecting Instruction Preferences (\method{})}
We start by defining the prompt selection problem in the pairwise preference optimization setting.
In this context, we present our proposed prompt-response-pair-based filtering method, which develops key descriptive metrics and their use in filtering training prompts.
Lastly, we describe how our method can be applied to self-instruction setups where synthetic prompts are generated from the model itself.

\subsection{Data Curation Problem}
\label{sec:problem}
The goal of  data curation  is to remove low-quality prompts that can negatively affect the general instruction following capability of the model.
Given a set of prompts $X=\{x\}$, we aim to find a subset $S \subseteq X$ to be used for fine-tuning a seed LLM $\mathcal{M}$.
We consider the preference optimization setting, with winning (chosen) and losing (rejected) response pairs $\{ y_w, y_l \}$ with  rewards $r(y_w | x) > r(y_l|x)$  for each prompt $x$.
The response pairs and their rewards can come from human preference data, 
or can be generated from the model itself $\mathcal{M}$ and then scored using an external reward model.
For the latter we use the "best-vs-worst" preference pairing method
\cite{pace2024west}, where $N$ responses are sampled, and the ones with highest and lowest rewards are the chosen and rejected, respectively: 
 \[\displaystyle
 \{y_i\}_{i=1}^N \sim \mathcal{M}(x) \quad \text{then} \quad
 \begin{cases}
 y_w &= \text{argmax}_{y_i} r(y_i | x) \\
 y_l &= \text{argmin}_{y_i} r(y_i | x)
 \end{cases} .
 \]
We also consider alternate pairing methods in \autoref{sec:ablation}.
We then use the preference data $\{x,y_w, y_l \}_{x \in S}$ for training the model $\mathcal{M}$.
Note that our focus is on filtering prompts entirely, not responses to those prompts.

\subsection{Hypothesis on Data Selection}
\label{sec:hypothesis}

Although preferences are extensively used to train state-of-the-art LLMs, there is limited research on identifying unhelpful training examples in this setting. We posit that analyzing the paired model responses to given input prompts can provide valuable insights into the quality of the prompts. Specifically, we test the following two hypotheses.

\para {Hypothesis 1: Low-quality prompts are likely to produce low-quality responses.}
Low-quality prompts - for example those that are unclear, ambiguous, or containing conflicting information - are likely to lead to noisy or inaccurate model responses. While those inaccurate responses can still be used as training targets in pairwise preference optimization, studies indicate that training on pairs with low-quality rejected responses might be sub-optimal. \citet{yasunaga2024alma} for example shows that pairing the best with random responses works well comparing to pairing the best with the worst one with lowest reward. This suggests a potential correlation of the quality of the rejected example with the alignment outcome. Additionally, several studies \citep{wu2024meta,zhao2024long,yuan2024following} have found a strong correlation between the length of responses, including rejected ones, and final performance. Therefore, we consider the reward $r(y_l|x)$ and length $\text{len}(y_l)$ of rejected responses as indicators of quality of the training prompts $x$, i.e. large values of either of these metrics relative to other examples indicate higher quality.

\para {Hypothesis 2. Low-quality prompts are likely to produce responses with larger variance}
Low quality prompts introduce uncertainty and ambiguity, leading to a broader range of interpretations. As the model or human generating the response might guess or fill in gaps in the prompt, this results in higher variance in responses. While some responses might align well with the intent, others may deviate significantly. A preliminary study in \citet{wu2024beta} finds low-gap pairs, where chosen and rejected responses are similar, are high-quality informative pairs, leading to better performing DPO models. %
We therefore consider the reward gap $r(y_w|x) - r(y_l|x)$ as another indicator of quality of a training prompt, i.e. small reward gaps suggest that the prompt has higher quality. 

\subsection{\method{} filtering}
\subsubsection{\method{} for existing training prompts}
\label{subsec:filtering}

Given the above hypotheses, we thus consider the following three metrics $m_k(x, y_w, y_l)$ that are based on the responses:
\begin{itemize}[itemsep=0mm,topsep=0mm,parsep=1mm]
    \item Rejected response reward: $m_1 = r(y_l|x)$
    \item Rejected response length: $m_2 = \text{len}(y_l)$ 
    \item Reward gap: $m_3 = r(y_w|x) - r(y_l|x)$
\end{itemize}
For each metric, we define threshold values that can be used for filtering.
For the first two metrics, higher values are desired so we choose a lower-bound threshold
\[\displaystyle
S = \{x \ | \ \tau_k \le m_k(x, y_w, y_l) \} .
\]
The last reward gap metric requires an upper threshold as we want small gaps.
Therefore we reduce the prompt selection problem to a threshold choice problem. %
To resolve this, 
we start with coordinate-wise experiments, analyzing model performance under various thresholds $\tau_k$ for individual metrics $m_k$ (details in \autoref{subsec:appex_additional_results}). 
Ultimately, we perform hyperparameter selection using all 3 parameters.

\subsubsection{Self-\method{} for synthetic prompts}
\label{sec:self_method}
Prompt curation by \method{} can also naturally be used to generate synthetic data. First, \method{}
is used to create a seed pool of high-quality prompts. Few-shot examples from this seed pool guide the model to generate training prompts, which can be further filtered by \method{}. We thus propose {\bf Self-\method{}}, a new approach to creating high-quality synthetic prompts:

\para{Step 1. Few-shot prompting with \method{} curated instructions}
We start with the set of prompts $S$ curated by our proposed method \method{} as described in \autoref{subsec:filtering}. To generate new prompts $S'$ we sample from our seed model $\mathcal{M}$ following Self-Instruct~\citep{wang2022self,honovich2022unnatural}.
For each new example we randomly select 8 prompts from $S$ and feed them as few-shot examples to the model $\mathcal{M}$ to generate a prompt with similar characteristics. We apply the exact processing steps in \citet{wang2022self} to new  prompts $S'$, such as removing similar prompts (ROUGE-L similarity with any existing instructions $<$ 0.7), and excluding those that contain certain keywords (e.g., image, picture, graph) that usually can not be processed by text-only LLMs.

\para{Step 2. Filtering with \method}
We further apply \method{} on top of the synthetically generated prompts $S'$ from the previous step, filtering out the self-instructions using the same threshold values as used before. Then the remaining subset $S''$ is used for training the seed model $\mathcal{M}$. %

Note we use \method{} filtering twice here, once in each step.
This is to ensure the quality of synthetic prompts. We also explore Self-\method{} using a smaller subset of $S$ as seed instructions in \autoref{sec:ablation} as part of our ablation studies.

\section{Experimental Setup}

We perform preference optimization %
using DPO, %
beginning with the Llama 3.1-8B-Instruct model as our seed model $\mathcal{M}$. We evaluate both the selection and creation of prompts, focusing on two categories: \textit{human-written} instructions and \textit{synthetically} generated instructions. Finally, we extend our evaluation of \method{} with the Llama 3.3-70B-Instruct model.

\subsection{Human-Written Prompts}
For human-written instructions, we specifically investigate two setups: human-written input prompts  1) paired with model-generated responses and annotated by a reward model; 2) with existing responses that have been annotated with human-assigned rewards. We use the WildChat and Helpsteer2 datasets, see statistics in Appendix \autoref{tab:data_stats}.

\subsubsection{WildChat dataset}
\para{Prompt Set} We start with a large pool of over 250k  human-written prompts from the WildChat~\cite{zhao2024wildchat} dataset. We exclude any non-English prompts based on WildChat annotations, and remove around 70k Midjourney-related instructions\footnote{They start with ``As a prompt generator for a generative AI called "Midjourney", you will create image prompts ...''.}, yielding 190k unique first-turn 
prompts.
These prompts are collected from real user interactions without human annotations, making them highly diverse. While there are many high-quality prompts, there are also a significant number of low-quality ones, such as nonsensical text or those lacking a clear question.

\para{Response Generation} Following \citet{yuan2024self,meng2024simpo,wu2024meta} we generate our chosen and rejected response pairs on the WildChat prompts using our seed model $\mathcal{M}$ to make our setup closer to the on-policy setting. 
We use best-vs-worst as described in \autoref{sec:problem},
generating $N$ responses for each prompt $x$ using $\mathcal{M}$ with sampling parameters of $T=0.8$, $top\_p = 0.95$.

\para{Reward Annotation} We then evaluate candidate responses using two different judges:
\begin{itemize}[itemsep=1mm,topsep=0mm,parsep=0mm]
    \item Reward Classifier: We used the ArmoRM reward model \citep{wang2024interpretable} to score each response.
    \item LLM-as-a-Judge \citep{zheng2023judging}: We prompt LLama 3.1-405B-Instruct using the prompt template outlined in \citet{yasunaga2024alma} to assign a score ranging from 0 to 10 for each response. For each response, we conduct 10 independent evaluations and use the average score as the final reward.
\end{itemize} 
The training example $(x, y_w, y_l)$ is selected by appointing the highest-reward one as $y_w$ and the lowest-reward one as $y_l$. 
For our primary experiments, we use the default value of \(N=64\). However, results for \(N=8, 16, 32\) are provided as part of our ablation studies in \autoref{tab:subset_abalation}, and we use \(N=32\) for the Llama 3.3-70B-Instruct experiments. We perform early stopping using a validation set of 470 examples: 253 valid set examples from \citet{li2023self} and 218 examples from the evol-test set of \citet{xu2023wizardlm}, with prompts that overlap with AlpacaEval2 removed.

\subsubsection{HelpSteer2 dataset} 
HelpSteer2 \cite{wang2024helpsteer2} consists of around 10k human-written prompts each with a response pair sampled from 10 different LLMs. Each response has human-annotated rewards of
helpfulness, correctness, coherence, complexity and verbosity on a Likert-5 scale. We use the aggregated reward with the recommended weighting [0.65, 0.8, 0.45, 0.55, 0.4].\footnote{\hyperlink{Llama3-70B-SteerLM-RM}{https://huggingface.co/nvidia/Llama3-70B-SteerLM-RM.}}
The main distinction from WildChat is that the rewards  come from human annotations instead of an external model.
We perform early stopping on the HelpSteer2 validation split, selecting checkpoints with the highest average response rewards determined by ArmoRM.

\subsection{Synthetic Prompts}

In this setup, we generate prompts from the seed model $\mathcal{M}$ itself for training instead of using human-written prompts. 
By varying the set of seed pool prompts used as few-shot examples, we collect two sets of training prompts:
\begin{itemize}[itemsep=1mm,topsep=0mm,parsep=0mm]
\item Self-Instruct: randomly select 8-shot examples from the unfiltered WildChat. 
\item Self-\method{}: randomly select 8-shot examples from high quality WildChat prompts filtered by \method{}.
\end{itemize}

In each case, we create 20k training prompts sampled with decoding parameters $T = 0.8$, $top\_p = 0.95$. The rest of the setup including response generations and DPO training is exactly the same as the WildChat setup where we use ArmoRM to construct response pairs $(y_w, y_l)$, and do early stopping on the same validation set of 470 examples.

\subsection{Baselines}
We compare our method with the existing methods below. For instruction-tuning data selection methods which handle a single (non-pairwise) response per prompt, we apply them to the chosen responses within the response pairs. Additional details on the implementation of each baseline are provided in Appendix \autoref{app:baseline_details}.

\subsubsection{Prompt-Based Filtering}

\para{InsTag Complexity} 
\citet{lu2023instag} leveraged ChatGPT to create semantic and intent-based tags, subsequently fine-tuning an LLM as a data tagger using these tags. They then used the tag counts as a measure of complexity. This is used to filter out prompts with fewer tags to enhance complexity.

\para{InsTag Diversity}
The InsTag Diversity filtering method \citep{lu2023instag} characterizes a dataset as more diverse when it includes a greater variety of unique tags, as annotated by the specified tagger. Using this approach, we greedily filter out data samples whose associated tags are already present in the selected dataset.

\para{LLM-as-Prompt-Judge}
 Employing LLMs as prompt quality judges has proven its efficacy in curating high-quality data \citep{chen2023alpagasus,dubey2024llama,liu2023makes}. We employ Llama 3.1-405B-Instruct to measure the quality of prompts on both a binary (useful/not useful) and pointwise scale (0-5). By sampling five Llama 3.1-405B-Instruct predictions per prompt and taking the average of LLM-as-Prompt-Judge predictions, we filter out less useful prompts by varying the cutoff thresholds.

\subsubsection{Prompt-and-Chosen-Response-Based Filtering}
\para{Perplexity} We compute perplexity (ppl) of the chosen response $y_w$ with the Llama 3.1-8B Instruct in a zero-shot manner as a filtering metric to curate training prompts. In particular, we retain examples with large $\text{ppl}(y_w|x)$ values, which may indicate the difficulty of the prompt.

\para{Instruction-Following Difficulty (IFD)}
\citet{li2023quantity} introduced the IFD to measure the model-specific difficulty of a data sample.
A lower IFD score indicates that this particular instruction-response pair is considered relatively easy for the language model to understand and follow without further training. %
We filter out examples with low IFD metric of a given pair of prompt $x$ and chosen response $y_w$.

\subsubsection{Chosen-and-Rejected-Response Based Filtering}

\para{Jaccard Similarity}
In addition to the reward gap between chosen and rejected responses, we explore  Jaccard similarity, defined as the number of overlapping words divided by the overall word counts, as an alternative similarity measurement. 
We thus filter out examples with low Jaccard similarity  scores (i.e. fewer overlapping words) between chosen and rejected response pairs.

\subsection{Training And Evaluation Setting}

Following the Instruct setup in \citet{meng2024simpo}, we utilize the DPO training approach with the off-the-shelf LLama 3.1-8B-Instruct and LLama 3.3-70B-Instruct models, leveraging the fairseq2 library \citep{balioglu2023fairseq2}. We use a batch size of $64$ and sweep over learning rates of $5e{-7}, 1e{-6}$ for the LLama 3.1-8B-Instruct model, and a learning rate of $1e{-6}$ with a batch size of $256$ for the LLama 3.3-70B-Instruct model. Both models are trained with a dropout rate of 0.0 and a $\beta$ value of 0.1 throughout the experiments. We conduct \method{} with various cutoff thresholds, e.g. at the 25\%, 50\% and 75\% percentile of each metric.

We primarily assess models' general instruction-following capabilities on three evaluation benchmarks: AlpacaEval2 \citep{alpaca_eval}, Arena-Hard \citep{arenahard2024} and WildBench \citep{lin2024wildbench}.  These benchmarks cover a wide range of natural yet challenging real-world user queries, and have been widely adopted by the research community.

\section{Experiment Results}

\begin{table*}[t!]
 \small
\setlength{\tabcolsep}{11pt}
  \centering
\caption{{\bf \method{} compared to existing filtering methods on WildChat with Llama 3.1-8B-Instruct.} 
\method{}, which selects only 4538 WildChat prompts for DPO training,  outperforms existing filtering methods on AlpacaEval2, Arena-Hard \& WildBench. DPO response pairs are constructed using ArmoRM to score responses.}
\begin{tabular}{lrcccc}
\toprule
& \multirow[b]{2}{4em}[-1.5mm]{\# Train examples} & \multicolumn{2}{c}{AlpacaEval2} & \multicolumn{1}{c}{Arena-Hard} & \multicolumn{1}{c}{WildBench}  \\
\cmidrule(lr){3-4} \cmidrule(lr){5-5}  \cmidrule(lr){6-6}
& & \multicolumn{1}{c}{LC Win} & \multicolumn{1}{c}{Win} & \multicolumn{1}{c}{Score}   & 
Score
\\
\midrule
\it{\textbf{Baseline}} \\
{Llama 3.1-8B-Instruct} (seed model)  & - & 20.9& 21.8  & 21.3  & 33.1\\
WildChat-20k DPO (no filtering) & 20000  & 48.4 & 45.9  & 37.9   & 41.5  \\
WildChat-20k DPO (best-vs-bottom-25\%) & 20000  &  48.2 & 45.9  & 40.7  & 44.5 \\
\midrule 
\multicolumn{3}{l}{\it{\textbf{Prompt-Based Filtering}}} \\
LLM-as-Prompt-Judge Binary  & 4299  &  45.5 & 41.0 & 42.0    & 43.3  \\
LLM-as-Prompt-Judge Pointwise  & 15963 & 47.4  & 47.4 & 40.7   & 45.2  \\
InsTag-Difficulty & 10000 & 46.3 & 39.0  &  39.0    & 42.4 \\
InsTag-Diversity  & 9952  &  40.1 & 41.1 &  40.4  & 43.4  \\
\midrule
\multicolumn{3}{l}{\it{\textbf{Prompt-and-Chosen-Response Based Filtering}}} \\
IFD on Prompt + Chosen Response & 9902 & 47.6 & 37.6 & 32.2  & 42.2  \\
ppl(Chosen Response) & 14851 & 45.6 & 45.5 &  40.8  & 43.4  \\
\midrule
\multicolumn{3}{l}{\it{\textbf{Chosen-Reject-Response Based Filtering}}} \\
Jaccard Similarity(Chosen, Rejected) & 9904 &  49.0 & 46.6  &  42.6    &  43.7	 \\ 
\method{} & 4538 &  \textbf{57.8} & \textbf{57.2}   & \textbf{43.1}  & \textbf{45.6}\\
\bottomrule
\end{tabular}
  \label{tab:alpaca_arena_wildbench_baseline}
\end{table*}

\begin{table*}[t!]
 \small
\setlength{\tabcolsep}{11pt}
  \centering
\caption{{\bf \method{} on WildChat with Llama 3.3-70B-Instruct.} 
\method{} outperforms no filtering on AlpacaEval2, Arena-Hard \& WildBench. DPO response pairs are constructed using ArmoRM to score responses.
}
\begin{tabular}{lrcccc}
\toprule
& \multirow[b]{2}{4em}[-1.5mm]{\# Train examples} & \multicolumn{2}{c}{AlpacaEval2} & \multicolumn{1}{c}{Arena-Hard} & \multicolumn{1}{c}{WildBench}  \\
\cmidrule(lr){3-4} \cmidrule(lr){5-5}  \cmidrule(lr){6-6}
& & \multicolumn{1}{c}{LC Win} & \multicolumn{1}{c}{Win} & \multicolumn{1}{c}{Score}   & 
Score
\\
\midrule
Llama 3.3-70B-Instruct (seed model) & - & 38.9 & 41.5 & 67.5 & 52.8 \\
WildChat-40k DPO (no filtering) & 40000  & 54.3 & 51.6 & 70.5 & 55.3  \\
\midrule 
\method{} & 17725 &  \textbf{67.7} & \textbf{73.2} & \textbf{82.9} & \textbf{58.8} \\
\bottomrule
\end{tabular}
  \label{tab:alpaca_arena_wildbench_llama3_3}
\end{table*}

Due to the large amount of unfiltered WildChat prompts, we first assess whether standard DPO training saturates as the size of the training prompts grows. As shown in Appendix \autoref{fig:wildchat_data_scaling}, the Armo Score on the valid set dramatically improves as we increase the size of training prompts, and begins to plateau afterwards. This shows growing the size of the training prompts arbitrarily does not bring additional gains, and hence quality control of the preference dataset could be important. %
We thus focus on 20k unique WildChat prompts, denoted as WildChat-20k for Llama3.1-8B-Instruct experiments, and 40k for Llama 3.3-70B-Instruct.

We report Alpaca-Eval2 Length-Controlled (LC) win rate, Arena-Hard score and WildBench WB-Score along with the number of training examples (after filtering if any)
using WildChat-20k
in \autoref{tab:alpaca_arena_wildbench_baseline}, on HelpSteer2 in \autoref{tab:alpaca_arena_wildbench_helpsteer2}, and on Self-Instruction data in \autoref{tab:alpaca_arena_wildbench_selfinstruct}. 
Existing filtering methods are provided in \autoref{tab:alpaca_arena_wildbench_baseline} as baseline comparisons. 
Further details, such as hyperparameters, are in Appendix  \autoref{tab:alpaca_arena_wildbench_wildchat_llmasjudge}  and \autoref{tab:full_results}.
Our findings lead to several key observations.

\begin{table*}[t!]
 \small
\setlength{\tabcolsep}{11pt}
  \centering
\caption{{\bf Weak to strong generation ability with \method{}.} \method{} on Llama 3.3-70B-Instruct by employing a smaller model Llama 3.1-8B-Instruct for filtering outperforms no filtering baseline, while underperforms using its own generations. }. 

\begin{tabular}{lrrcccc}
\toprule
& \multirow[b]{2}{4em}[-1.5mm]{Filter Model} & \multirow[b]{2}{4em}[-1.5mm]{\# Train examples} &\multicolumn{2}{c}{AlpacaEval2} & \multicolumn{1}{c}{Arena-Hard} & \multicolumn{1}{c}{WildBench}  \\
\cmidrule(lr){4-5} \cmidrule(lr){6-6}  \cmidrule(lr){7-7}
 Seed Model & & & \multicolumn{1}{c}{LC Win} & \multicolumn{1}{c}{Win} & \multicolumn{1}{c}{Score}   & Score \\

\midrule
Llama 3.3-70B-Instruct & - (no filtering) & 40000  & 54.3 & 51.6 & 70.5 & 55.3  \\
\midrule
Llama 3.3-70B-Instruct & Llama 3.1-8B-Instruct & 18184 & 64.5 & 69.2 & 76.7 & 58.6\\
Llama 3.3-70B-Instruct & Llama 3.3-70B-Instruct & 17725 & \textbf{67.7} & \textbf{73.2} & \textbf{82.9} & \textbf{58.8}\\
\bottomrule
\end{tabular}
  \label{tab:weak_to_strong}
  \vspace{-2mm}
\end{table*}
\para{When filtering human-written instructions, \method{} achieves the best performance on both human-scored and model-scored preference datasets.} On the WildChat dataset where pairs are annotated by the ArmoRM model, we conduct \method{} with various cutoff thresholds, at the 25\%, 50\% and 75\% percentile of each metric. Our best model is trained on examples with rejected length larger than the 50\% percentile of all rejected lengths, and rejected rewards  larger than the 50\% percentile of all rejected rewards, and reward gap smaller than the 50\% percentile. \autoref{tab:alpaca_arena_wildbench_baseline} shows that \method{} significantly improves LC win rate from the LLama3.1-8B-Instruct DPO baseline without filtering from 48.4\% to 57.8\% by filtering out 77\% training examples, surpassing GPT-4 Omni (05/13) on AlpacaEval2. Similarly, \method{} scores the highest on Arena-Hard (43.1) compared to LLM-as-Prompt-Judge filtering (42.0), Jaccard Similarity (42.6), and the no filtering baseline (37.9). \method{} also achieves the highest WB-score on WildBench (45.6) compared to other filtering and no filtering baselines (41.5). %
As shown in Appendix \autoref{tab:alpaca_arena_wildbench_wildchat_llmasjudge} using LLM-as-a-Judge annotated rewards, \method{} also performs well. Finally, \autoref{tab:alpaca_arena_wildbench_helpsteer2} demonstrates \method{} is equally effective on HelpSteer2 where preference pairs are determined by human annotators, achieving the highest scores across all 3 evaluation benchmarks as compared to the baselines (no filtering and LLM-as-Prompt-Judge filtering).

\para{\method{} scales to different and larger models} 
We also tried \method{} on a different base LLM --  from the Llama 3.3  family rather than 3.1, and of a larger scale, 70B rather than 8B.
As shown in \autoref{tab:alpaca_arena_wildbench_llama3_3}, \method{} also works on this larger model. Filtering dramatically boosts Llama 3.3-70B-Instruct DPO trained models, with AlpacaEval2 LC win rate improved from 54.3\% to 67.7\%, Arena Hard from 70.5 to 82.9 and WildBench from 55.3 to 58.8, surpassing SOTA models as shown in \autoref{tab:alpaca_arena_wildbench_wildchat_sota}. The prompt filtering threshold we applied to the 70B model was the same as in Llama 3.1-8B-Instruct + \method{} (see Appendix \autoref{tab:full_results}).%

\para{Weak-to-strong generalizability of \method{} } 
To explore potential weak-to-strong generalizability \citep{li2024superfiltering} of our method, we employ a smaller and weaker model, Llama 3.1-8B-Instruct, to filter data for a larger and more powerful LLM, Llama 3.3-70B-Instruct. As illustrated in \autoref{tab:weak_to_strong}, while the filtering capability of Llama 3.1-8B-Instruct is not as powerful as that of Llama 3.3-70B-Instruct, it still offers significant improvements over baseline no filtering. This showcases the weak-to-strong generation capabilities of our \method{}, demonstrating that leveraging a smaller model to assist a larger one in data filtering is a computationally efficient strategy.

\para{Existing filtering methods derived from supervised-finetuning do not work as well on preference datasets} As demonstrated in \autoref{tab:alpaca_arena_wildbench_baseline}, compared to the baseline WildChat-20k DPO (no filtering) trained on WildChat 20k prompts without any filtering, existing prompt-based filtering methods such as InsTag-Difficulty,  InsTag-Diversity or LLM-as-Prompt-Judge filtering methods all lead to lower win rates on Alpaca-Eval2. LLM-as-Prompt-Judge, while outperforming certain filtering methods such as InsTag, achieves marginal gains compared to no filtering even though they are facilitated by querying a poweful LLM, Llama 3.1-405B-Instruct. Out of all the alternative methods tried,
Jaccard Similarity based filtering that takes into account response pairs for filtering achieves relatively the  highest scores across the 3 benchmarks, indicating that filtering that only takes into account prompts or chosen responses does not generalize well to the pairwise preference case.

\para{The Self-\method{} method to generate synthetic data outperforms Self-Instruct data.}
As shown in \autoref{tab:alpaca_arena_wildbench_selfinstruct},  Self-\method{} yields better alignment results across all 3 evaluation benchmarks as compared to those trained on Self-Instruct data. In particular, win rate improves from 49.1\% to 60.2\% on AlpacaEval2, and from 38.5\% to 42.1\% on Arena-Hard. 
This result implies that our method generates better quality instructions than generating via few-shot examples from unfiltered prompts as in Self-Instruct.

\para{Self-\method{} synthetic data %
outperforms human-written instructions}
In \autoref{tab:alpaca_arena_wildbench_selfinstruct}, models trained on %
synthetic prompts  %
outperform those trained on 20k human-written WildChat prompts.
Applying Self-\method{} few-shot generation {\em without post-filtering} gives an equal amount of 20k prompts, but still increases  the AlpacaEval2 LC win rate from 48.4\% to 53.6\%, Arena-Hard win rate from 37.9\% to 43.7\% and WB-Score on WildBench from 41.5 to 44.8. This further illustrates the importance of training on high-quality instructions. When applying the full Self-\method{} method with post-filtering results are further improved, for example achieving the best AlpacaEval2 LC win rate of 60.2\%.

\para{\method{} seed data selection  and \method{} post-filtering are both important for generating Self-\method{} synthetic data}
In \autoref{tab:alpaca_arena_wildbench_selfinstruct},  we perform ablations on Self-\method{}.  We try: (i) using~\method{} to select high quality few-shot examples  but not for curating the resulting generations (post-filtering); (ii) applying standard (Self-Instruct) few-shot generation, but then applying  \method{} post-filtering;  
or (iii) applying \method{} to both few-shot generation and post-filtering  (our default method). 
We find that both components of our full method are important yielding the best results,  with method (i) outperforming Self-Instruct, and method (ii) performing better than (i), but worse than our full method (iii).

\begin{table*}[t!]
 \small
\setlength{\tabcolsep}{11pt}
  \centering
\caption{
{\bf \method{} on HelpSteer2 with Llama 3.1-8B-Instruct.}
Applying \method{} to DPO models trained on HelpSteer2 outperforms the baseline of no filtering as well as using the Llama 3.1-405B-Instruct model as a pointwise prompt quality judge.}
\begin{tabular}{lrcccc}
\toprule
& \multirow[b]{2}{4em}[-1.5mm]{\# Train examples} &\multicolumn{2}{c}{AlpacaEval2} & \multicolumn{1}{c}{Arena-Hard} & \multicolumn{1}{c}{WildBench}  \\
\cmidrule(lr){3-4} \cmidrule(lr){5-5}  \cmidrule(lr){6-6}
 HelpSteer2 &  & \multicolumn{1}{c}{LC Win} & \multicolumn{1}{c}{Win} & \multicolumn{1}{c}{Score}   & Score \\
\midrule
{Llama 3.1-8B-Instruct} (seed model)  & - & 20.9& 21.8 & 21.3  &   33.1\\
HelpSteer2 DPO (no filtering) & 10161 & 25.2 & 23.1  & 26.8  &  37.1\\
\midrule
LLM-as-Prompt-Judge filtering & 5376  & 27.8  & 25.7  & 29.5  & 37.2  \\
\method{}  & 5081  & \textbf{34.6} & \textbf{32.8}  &  \textbf{35.0}  &  \textbf{39.5}	 \\
\bottomrule
\end{tabular}
  \label{tab:alpaca_arena_wildbench_helpsteer2}
\end{table*}

\begin{table*}[t!]
 \small
\setlength{\tabcolsep}{11pt}
  \centering
\caption{{\bf Self-\method{} for generating high-quality synthetic instructions}. 
Self-\method{} creates prompts using few-shot samples from high-quality prompts curated by \method{}, whereas Self-Instruct uses few-shots from unfiltered WildChat prompts. Applying \method{} filtering {\em after} generation is also important, and achieves the best  results, significantly outperforming  Self-Instruct data.}
\begin{tabular}{lrrcccc}
\toprule
& \multirow[b]{2}{4em}[-1.5mm]{Post-Filtering} & \multirow[b]{2}{4em}[-1.5mm]{\# Train examples} &\multicolumn{2}{c}{AlpacaEval2} & \multicolumn{1}{c}{Arena-Hard} & \multicolumn{1}{c}{WildBench}  \\
\cmidrule(lr){4-5} \cmidrule(lr){6-6}  \cmidrule(lr){7-7}
{Train Prompts} &    & & \multicolumn{1}{c}{LC Win} & \multicolumn{1}{c}{Win} & \multicolumn{1}{c}{Score}   & Score \\
\midrule
WildChat-20k & None & 20000 & 48.4 & 45.9  & 37.9   & 41.5  \\
WildChat-20k & \method{}& 4538 &  57.8 & 57.2   & 43.1  & \textbf{45.6}\\
\midrule
Self-Instruct & None  & 20000 & 49.1 & 46.9  &  38.5 &  40.0 \\
Self-\method{} (without post-filtering) & None & 20000 & 53.6 & 56.1  &  \textbf{43.7}   & 44.8  \\
Self-Instruct with \method{}~post-filtering & \method{} & 16261 & 58.3 & 53.2  &   40.9& 44.1  \\
Self-\method{}  & \method{} & 18812 & \textbf{60.2} & \textbf{61.1}  &   42.1  &  42.5 \\
\bottomrule
\end{tabular}
  \label{tab:alpaca_arena_wildbench_selfinstruct}
  \vspace{-2mm}
\end{table*}

\section{Understanding why \method{} works}

\subsection{Filtering prompts with low quality responses}

To understand  what instructions are filtered out, we first visualize instructions with low quality rejected responses (as measured by  low reward and short lengths) by comparing the t-SNE plots of unfiltered and filtered instructions (shown in Appendix \autoref{fig:wildchat_unfiltere_vs_7k}). We investigated a few clusters present in that t-SNE plot of unfiltered prompts that are missing from the t-SNE plot of filtered ones on the right-hand-side. %
We find that instructions from those clusters being filtered out from the training set are either obscure, non-sensical, or they fail to elicit meaningful responses from the model, leading to lower-quality rejected responses. Such instructions can be caught by measuring the rewards and lengths of the rejected responses, with supporting evidence given in Appendix \autoref{tab:wildchat_unfiltere_vs_7k_by_cluster}.

Next, we employ GPT-4 and LLama3.1-405B-Instruct to evaluate first 10,000 prompts from WildChat. Focusing solely on the instructions (excluding responses) provided in WildChat, the model is tasked with scoring each prompt on a scale from 1 to 5. A score of 1 represents the most helpful prompt, while a score of 5 indicates the lowest quality. The evaluation prompt is provided in Appendix \autoref{fig:gpt4_eval_prompt}. Manual review revealed that prompts assigned scores of 4 and 5 were of very low quality, while those scored  3 were moderately acceptable, albeit with some quality issues still present. Notably, GPT-4 and LLama3.1-405B-Instruct occasionally assigned scores of 2 or 3 to a prompt of low quality. 
\autoref{tab:gpt4_annotate_quality_safety} illustrates the prevalence of low-quality examples (with score 4 or 5 by both GPT-4 and LLama3.1-405B-Instruct) after applying various filtering methods. We observe that filtering based on the reward and length of the rejected response is the most effective way to ensure prompt quality, compared to other methods tried. By combining those rejected response quality metrics with the reward gap, \method{} reduced percentage of noisy prompts from 22.9\% to 8.9\%. This supports our hypothesis that very low-quality prompts, such as those in WildChat that consist of incomplete snippets from movies, stories, or code (see sample rejected instructions in Appendix \autoref{tab:wildchat_unfiltere_vs_7k_by_cluster} and \autoref{tab:wildchat_7k_vs_5k_by_cluster}), often result in poor rejected responses when sampled several times. By leveraging the quality of rejected responses as a filtering criterion, we can efficiently eliminate these extremely noisy prompts.

Furthermore, we employ GPT-4 and LLama3.1-405B-Instruct to respond to each WildChat prompt three times. If any response declines to answer due to safety concerns, we categorize those prompts as unsafe. It's important to note that with this method, the model sometimes assigns high quality scores to prompts that are borderline unsafe. By examining the reward and the length of rejected responses, we observe \method{} is also an effective approach to filter out these unsafe prompts. This approach is grounded in the observation that rejected responses when dealing with unsafe instructions are typically short and have low reward scores. %

\subsection{Filtering prompts with larger response variance}
Similarly, we visualize instructions that are filtered out by measuring the reward gap between chosen and rejected responses in Appendix \autoref{fig:wildchat_7k_vs_5k}, and further expand some representative groups of filtered instructions in Appendix \autoref{tab:wildchat_7k_vs_5k_by_cluster}. In particular, instructions that cover specialized domains such as coding, software, and other technical questions often require precise details, well-defined objectives or targeted solutions. In those cases,  a lack of specificity in the instructions might lead to more variable responses. As shown in \autoref{tab:wildchat_7k_vs_5k_by_cluster}, instructions with larger reward gap are not necessarily low-quality, however we hypothesize that the combination of lack of specificity in the instruction and larger difference in the response pair make them less helpful in improving the model during preference optimization.

\begin{table}[ht!]
\small
    \centering
    \caption{ {\bf Effectiveness of Filters on Prompt Quality and Safety}: we compare the number of noisy and potentially unsafe (as judged by GPT4) WildChat instructions (out of 20k) filtered by various filtering methods.}
    \begin{tabular}{p{2.5cm}|r|r}
    \toprule
Filtering Methods & \thead{\% of low-quality\\prompts $\downarrow$} & \thead{\% of unsafe\\prompts $\downarrow$} \\
\midrule
Unfiltered Data & 22.9\% & 12.27\% \\
Reject Reward & 10.4\% & 0.04\%\\
Reject Length & 13.9\% & 0.02\% \\
Reward Gap & 17.7\% & 8.07\%\\
\method{} & 8.9\% & 0.00\%\\
         \bottomrule
    \end{tabular}
    \label{tab:gpt4_annotate_quality_safety}
\end{table}

\section{Conclusion}
This work introduces \methodlong{} (\method{}), a method for improving preference data quality by measuring the rejected response quality and the reward gap between the chosen and rejected response pair. Filtering instructions using \method{} remarkably improves model alignment results on both human-written and synthetic instructions, and for different reward signals. In addition, we show that Self-\method{}, synthetic instructions generated by few-shot prompts curated by \method, outperforms organic user instructions and the standard Self-Instruct method,  
achieving the highest AlpacaEval2 win rate in our experiments.

\section*{Impact Statement}
This work demonstrates the possibility of dramatically improving LLMs by identifying and producing high-quality training data. Studying how filtering criteria affect outputs will continue to be important for LLM training. 
While we have primarily focused on preference optimization,  the \method{} approach is general and can potentially work for any training scheme, e.g. other RL training techniques -- which future work should explore.

For such models, safety will also be crucial, and future work should additionally address this aspect. In our  experiments, the  reward is not explicitly constrained by safety-related criteria. 
Therefore, a clear further avenue of study is to conduct safety evaluations -- and to explore safety filtering using our methods, with reward models built exclusively for safety in existing systems \citep{touvron2023llama2}.

Given that we have shown that \method{} can filter potentially unsafe prompts, this could mean in the best case that the safety of the model could potentially improve after filtering as well, with \method{} being able to catch and mitigate more challenging safety situations that earlier iterations cannot.
From a broader perspective, this work could pave the way for methods that produce higher-quality training instructions, that are also potentially safer than organic user instructions in the wild. 

\bibliography{sample}
\bibliographystyle{icml2025}

\newpage
\appendix
\onecolumn

\section{Appendix}

\subsection{More Details on Experiment Setup}

Our experiment setups are summarized in \autoref{tab:data_stats}. Specifically, we apply \method{} to multiple popular instruction-following datasets as well as our own synthetic data, with reward annotated from various sources (human/reward classifier/LLM-as-a-Judge), indicating the generalizability of our \method{} method.

\begin{table*}[ht!]
 \small
 \caption{{\bf Preference Dataset Statistics} used for training in our experiments.}
\setlength{\tabcolsep}{11pt}
  \centering
\begin{tabular}{lrrrrr}
\toprule
  &               & Human &  & Reward  \\
  &  \#Prompts    &  Written &   \#Responses &  Annotator  & Valid Set ( \# Examples)\\
\midrule
WildChat-turn1 20k & 20,000 & Yes & 8,16,32,64 & ArmoRM  &  Humpback + Evol-Instruct (470) \\
WildChat-turn1 20k & 20,000 & Yes & 64 & LLM-as-a-Judge  &  Humpback + Evol-Instruct (470) \\
HelpSteer2 & 10,161  & Yes &  2 & Human  & HelpSteer2 valid (519) \\
Self-Instruct  & 20,000 & No & 64 & ArmoRM & Humpback + Evol-Instruct (470) \\
Self-\method{} & 20,000 & No & 64 & ArmoRM & Humpback + Evol-Instruct (470) \\
\bottomrule
\end{tabular}
  \label{tab:data_stats}
\end{table*}

\begin{figure}[ht!]
    \caption{{\bf Results on DPO Training with Varying WildChat Data Sizes.} Using different sizes of WildChat data for DPO training on LLaMA 3.1-8B-Instruct, the performance, measured by Armo rewards on the validation set, gradually saturates as the data size increases.}
    \centering
    \includegraphics[width=0.45\textwidth]{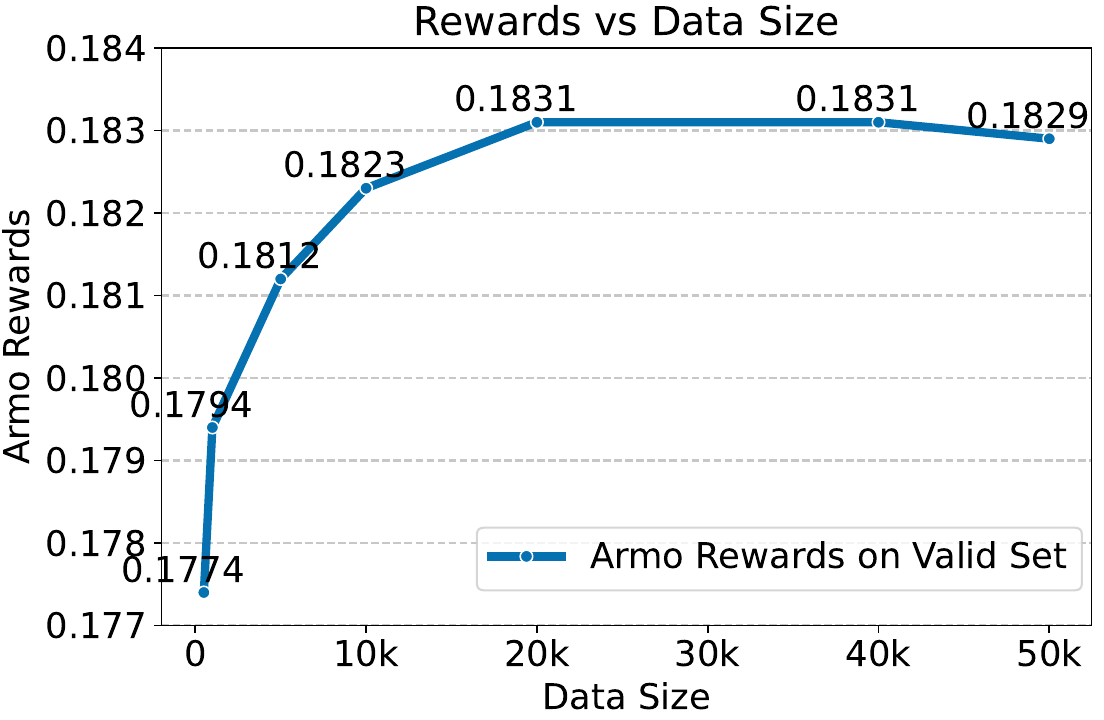}
    \label{fig:wildchat_data_scaling}
\end{figure}

We report the model performance on valid set when varying the number of training WildChat prompts in \autoref{fig:wildchat_data_scaling}. Model training improves significantly as training data size grows to 20k and then begin to saturates afterwards, therefore our main experiments are based on those 20k WildChat prompts.

We primarily assess our models' general instruction-following capabilities using three popular evaluation benchmarks: AlpacaEval-2 \citep{alpaca_eval}, Arena-Hard \citep{arenahard2024} and WildBench \citep{lin2024wildbench}. 
AplacaEval-2 consists of 805 prompts sampled from 5 datasets. Arena-Hard contains 500 challenging user queries sourced from Chatbot Arena and has the highest correlation and separability of models commpared to Chatbot Arena among popular open-ended LLM benchmarks \citep{arenahard2024}. WildBench is built from a set of 1024 significantly harder, challenging queries carefully curated from the WildChat project \citep{zhao2024wildchat} to ensure diversity and complexity. The automatic evaluation of WildBench involves task-specific checklists that guide LLM judges in generating reliable and consistent judgments which demonstrate significantly high correlation with human judgments. We report the WB-Score for individual scoring.

\subsection{Additional Results}
\label{subsec:appex_additional_results}

\begin{table*}[ht!]
 \small
\setlength{\tabcolsep}{11pt}
  \centering
\begin{tabular}{lrcccc}
\toprule
& \multirow[b]{2}{4em}[-1.5mm]{\# Train examples} &\multicolumn{2}{c}{AlpacaEval2} & \multicolumn{1}{c}{Arena-Hard} & \multicolumn{1}{c}{WildBench}  \\
\cmidrule(lr){3-4} \cmidrule(lr){5-5}  \cmidrule(lr){6-6}
 WildChat with LLM-as-a-Judge as reward annotator &  & \multicolumn{1}{c}{LC Win} & \multicolumn{1}{c}{Win} & \multicolumn{1}{c}{Score}   & Score \\
\midrule
{Llama 3.1-8B-Instruct} (seed model)  & - & 20.9& 21.8 & 21.3  &   33.1\\
Standard DPO (no filtering) & 16837 & 40.1 & 44.9  & 41.1  & 42.5 \\
\midrule
\method{} & 5999  & 44.3  & 48.8  & 42.5  &  43.9 \\
\bottomrule
\end{tabular}
\caption{{\bf \method{} compared to baselines on WildChat using LLM-as-a-Judge as the reward annotator.}. 
We report results on AlpacaEval2, Arena-Hard and WildBench of various models trained using DPO on the WildChat Dataset. \method{} outperforms the baseline of LLM-as-judge as the reward annotator.}
\label{tab:alpaca_arena_wildbench_wildchat_llmasjudge}
\end{table*}

\paragraph{LLM-as-a-Judge As Reward Annotators} We explore LLM-as-a-Judge as alternative reward annotator apart from the reward model ArmoRM and human reward annotations, and use a LLama3.1-405B-Instruct zero-shot to judge the quality of each individual response and uses its prediction to construct response pairs. For each response, we conduct 10 independent evaluations and calculate the average score to determine the final reward score. We report AlpacaEval2, Arena-Hard and WildBench results on WildChat DPO models in \autoref{tab:alpaca_arena_wildbench_wildchat_llmasjudge}. Similar to the observation from \autoref{tab:alpaca_arena_wildbench_baseline}, \method{} by filtering based on LLM-as-a-Judge predictions outperforms no filtering. 

\paragraph{Data Scaling with \method{}}  We further scale up \method{} by growing the training data size after filtering to 20k, and achieves AlpacaEval2 LC win rate of 58.49\% as shown in \autoref{fig:data_scaling}. While the effective training size scales from 4538 to 20k, the actual performance gain only increase slightly, suggesting that training with Llama 3.1-8B-Instruct on existing WildChat prompts saturates, even under \method{}.

\paragraph{\method{} filtering thresholds}
We report the filtering thresholds of the best checkpoints in our experiments in \autoref{tab:full_results}.

\paragraph{Full Evaluation Results} We include full WildChat evaluation results on AlpacaEval2 and Arena-Hard in \autoref{tab:alpaca_arena_wildchat} and on WildBench in \autoref{tab:wildbench_wildchat}, with average response lengths, confidence intervals as well as finegrained results on subtasks. Full evaluation results on models trained on HelpSteer2 are presented in \autoref{tab:alpaca_arena_helpsteer2} and \autoref{tab:wildbench_eval_helpsteer}. In addition, full evaluation results on Self-\method{} are included in \autoref{tab:alpaca_arena_selfinstruct} and \autoref{tab:wildbench_eval_self_instr}.

\paragraph{Coordinate-wise Filtering results.} We conduct extensive experiments  by applying filtering to each individual metric: reward on chosen or rejected response, lengths of chosen or rejected response, reward gap, average reward of all responses, etc. Results on valid set performances by applying various filtering metrics to WildChat task are included in \autoref{table:wildchat_coordinate_wise}, and HelpSteer2 in \autoref{table:helpsteer_coordinate_wise}. Both highlight strong performance boost by filtering based on rejected reward, rejected length and reward gap.

\begin{sidewaystable}
    \centering
    \small
    \begin{tabular}{>{\centering\arraybackslash}p{2.5cm}p{1cm}p{1cm}p{1cm}p{1.5cm}p{1.8cm}>{\centering\arraybackslash}p{5cm}ccp{1.5cm}p{1cm}}
        \toprule
        & \multicolumn{6}{c}{} &\multicolumn{2}{c}{AlpacaEval2} & \multicolumn{2}{c}{} \\
        \cmidrule(lr){8-9} \\
        Data & \# Train Examples & Human Written & \# Responses & Reward Annotator&Seed Model &  Filtering Metrics &  LC Win & Win & Arena-Hard & Wildbench \\
        \hline
        - & - & -& -& - & Llama 3.1-8B-Instruct & - & 20.9 & 21.8&21.3 & 33.1 \\
        - & - & -& -& - & Llama 3.3-70B-Instruct & - & 38.9 & 41.5& & 52.8 \\
        \hline
        Wildchat 20k & 20k & Yes & 64 & ArmoRM & Llama 3.1-8B-Instruct  & No & 48.37 & 45.87& 37.9&41.5 \\
        Wildchat 20k & 6762 & Yes & 64 & ArmoRM & Llama 3.1-8B-Instruct & Rejected Length $\geq$ 1878, Rejected Armo $\geq$ 0.126 & 57.1 & 52.9 & 42.3 & 45.5 \\
         Wildchat 20k & 4538 & Yes & 64 & ArmoRM & Llama 3.1-8B-Instruct  & Rejected Length $\geq$ 1878, Rejected Armo $\geq$ 0.126, Reward Gap $>$ 0.042 &57.8 & 57.2 &43.1 & 45.6\\ 
         \hline
         Synthetic (few shot: Wildchat 20k) & 20k & No & 64 &  ArmoRM & Llama 3.1-8B-Instruct & No & 49.1& 46.9& 38.5& 41.0\\
         Synthetic (few shot: Wildchat 20k) & 16k & No & 64 & ArmoRM & Llama 3.1-8B-Instruct & Rejected Length $\geq$ 1878, Rejected Armo $\geq$ 0.126 & 58.3 & 53.2 & 40.9 & 44.1\\
         Synthetic (few shot: Wildchat filtered 4538 examples)& 20k & No & 64 & ArmoRM & Llama 3.1-8B-Instruct & No & 53.6 & 56.1 & 43.7 & 44.8\\
         Synthetic (few shot: Wildchat filtered 4538 examples) & 18812 & No &64& ArmoRM & Llama 3.1-8B-Instruct & Rejected Length $\geq$ 1878, Rejected Armo $\geq$ 0.126 & 60.2 & 61.1 & 42.1 & 42.5\\
         \hline
         Wildchat 20k & 16.8k & Yes & 64 & LLM-as-a-Judge & Llama 3.1-8B-Instruct & No & 40.1 & 44.9 & 41.1 & 42.5 \\
         Wildchat 20k & 5999 & Yes & 64 & LLM-as-a-Judge & Llama 3.1-8B-Instruct & Rejected LLM-as-a-Judge Reward $\ge$ 8, Rejected Length $\ge$ 1399, Reward Gap $\le$ 1 & 44.3 & 48.8 & 42.5&43.9\\
         \hline
         HelpSteer & 10k & Yes & 64 & Human& Llama 3.1-8B-Instruct & No & 25.2 & 23.1 & 26.8& 37.1 \\
         HelpSteer & 5081 & Yes& 64 & Human& Llama 3.1-8B-Instruct & Rejected Length $\ge$ 1303 & 34.6& 32.8 & 35.0& 39.5\\
         \hline
         Wildchat 40k & 40k & Yes & 32 & ArmoRM & Llama 3.3-70B-Instruct & No & 54.3 & 51.6 & 70.5 & 55.3  \\
         Wildchat 40k & 17.7k & Yes & 32 & ArmoRM & Llama 3.3-70B-Instruct & Rejected Length $>$ 1878, Rejected Armo $>$ 0.126, GAP $>$ 0.042 &67.7 & 73.2 & 82.9 & 58.8 \\
        \bottomrule
    \end{tabular}
    \caption{{\bf Full Results} details on number of training examples, choice of reward models, seed models, filtering metrics and thresholds chosen as well as final outcomes across 3 evaluation benchmarks.}
\label{tab:full_results}
\end{sidewaystable}

\begin{table*}[t!]
 \small
 \caption{{\bf Full AlpacaEval2 \& Arena-Hard Results on WildChat}: we compare performances of SOTA models on AlpacaEval2 win rates and Arena-Hard scores as well as DPO models trained on the WildChat-20k dataset using various filtering methods.}\label{tab:alpaca_arena_wildchat}
\setlength{\tabcolsep}{8pt}
  \centering
\begin{tabular}{lrrrrrrrc}
\toprule
&  \multirow[b]{2}{4em}[-1.5mm]{\# Train examples}&\multicolumn{3}{c}{AlpacaEval2} & \multicolumn{3}{c}{Arena-Hard}  \\
\cmidrule(lr){3-5} \cmidrule(lr){6-8}
\emph{Standard models} &  & \multicolumn{1}{c}{LC Win} & \multicolumn{1}{c}{Win} & \multicolumn{1}{c}{Len} & \multicolumn{1}{c}{Score}   & \multicolumn{1}{c}{95\% CI} & \multicolumn{1}{c}{Len} \\
\midrule
GPT-4 Omni (05/13)  & -  & 57.5 & 51.3 & 1873 & 74.9  & (-2.5, 1.9) & 668 \\
GPT-4 Turbo (04/09)  & -  & 55.0 & 46.1 & 1802   & 82.6 & (-1.6, 1.8) & 662  \\
Gpt-4-0613  & -  & 55.0 & 46.1 & 1802   & 37.9 & (-2.8, 2.4) & 354 \\
Llama 3.1-405B-Instruct & -  & 39.3 & 39.1 & 1988 & 67.1  & (-2.2, 2.8) & 658 \\
{Llama 3.1-70B-Instruct}  &  -  & 38.1  & 39.1 &  2044   & 69.3 & (-2.5, 2.5) & 658  \\
\midrule
\it{\textbf{Baseline}} \\
{Llama 3.1-8B-Instruct}  & - & 20.9& 21.8 &  2184 & 21.3 & (-1.9, 2.2) &  861\\
WildChat-20k DPO (no filtering) & 20000  & 48.4 & 45.9 & 2134  & 37.9 & (-2.0, 2.2)  & 622 \\
WildChat-20k DPO (best-vs-bottom-25\%) & 20000  &  48.2 & 45.9 & 1971 & 40.7 & (-2.1, 1.9) & 741 \\
\midrule 
\multicolumn{3}{l}{\it{\textbf{Prompt-Based Filtering}}} \\
Jaccard Similarity & 9904 &  49.0 & 46.6 & 1978 &  42.6  & (-2.4, 2.3) & 632\\
LLM-as-Prompt-Judge Binary  & 4299  &  45.5 & 41.0 & 1859 & 42.0 &  (-1.4, 1.7)   & 597 \\
LLM-as-Prompt-Judge Pointwise  & 15963 & 47.4  & 47.4 & 2056 & 40.7  & (-2.0, 2.2) & 701 \\
InsTag-Difficulty & 10000 & 46.3 & 39.0 & 1752 &  39.0 &  (-2.2, 2.3)  & 602 \\
InsTag-Diversity  & 9952  &  40.1 & 41.1& 1903 &  40.4  & (-2.4, 2.8) & 579 \\
\midrule
\multicolumn{3}{l}{\it{\textbf{Prompt-and-Chosen-Response-Based Filtering}}} \\
IFD on Prompt + Chosen Response & 9902 & 47.6 & 37.6 & 1655 & 32.2 & (-1.7, 2.5)  & 533 \\
ppl(Chosen Response) & 14851 & 45.6 & 45.5 & 1930 &  40.8 & (-2.3, 1.7) & 582 \\
\midrule
\multicolumn{3}{l}{\it{\textbf{Chosen-Rejected-Response Based Filtering}}} \\
LLM-as-Prompt-Judge Pointwise  &  15963 & 47.4 & 47.4 & 2056 & 40.7 & (-2.0, 2.2) & 701 \\
\method{} & 4538 &  57.8 & 57.2 & 2048  & 43.1  & (-1.5, 1.8)  & 638 \\
\bottomrule
\end{tabular}
\end{table*}

\begin{table*}[t!]
 \small
 \caption{{\bf Full WildBench Results on WildChat}: we compare performances of SOTA models on WildBench as well as DPO models trained on the WildChat-20k dataset using various filtering methods.}
  \centering
\begin{tabular}{lrrrrrrc}
\toprule
&\thead{WB-Score} & \multicolumn{5}{c}{WB-Score: Task-specific}  \\
\cmidrule(lr){2-2} \cmidrule(lr){3-7}
\emph{Standard models} & \thead{Weighted \\ average} & \multicolumn{1}{c}{Creative} & \thead{Planning  \& \\ Reasoning} & \thead{Math \& \\ Data Analysis}   & \thead{Information} & \thead{Coding \& \\ Debugging} 
\\
\midrule
GPT-4 Omni (05/13) & 59.3	& 59.1	& 60.2	& 57.3	& 58.6	& 60.5 \\
GPT-4 Turbo (04/09)  & 55.2 &	58.7 &	56.2	& 51.0	& 57.2	& 55.1  \\
Gemini-1.5-pro & 53.0 &	55.1	& 53.7 &	48.6	& 52.2	& 55.2 \\
{Llama3-70B-Instruct}  & 47.8 &	54.3 &	50.1	& 42.1	& 52.3	& 44.7  \\
\midrule
\it{\textbf{Baseline}} \\
{Llama 3.1-8B-Instruct}  & 33.1 & 45.0 & 37.0 & 23.9 & 37.4 & 29.3\\
WildChat-20k DPO (no filtering) & 41.5 & 51.8 & 44.2 & 32.2 & 50.0 & 37.1\\
WildChat-20k DPO (best-vs-bottom-25\%) & 44.5 & 53.9 & 47.4 & 35.8 & 50.4 & 41.4\\
\midrule 
\it{\textbf{Prompt-Based-Filtering}} \\
Jaccard Similarity &  43.7 & 54.2 & 46.9 & 34.3 & 49.5 & 40.5\\
LLM-as-Prompt-Judge Binary  & 43.3 & 53.9 & 46.6 & 35.8 & 48.5 & 38.6 \\
LLM-as-Prompt-Judge Pointwise  & 45.2 & 55.6 & 48.0 & 37.1 & 51.6 & 40.9\\
 InsTag-Difficulty & 42.4 & 52.7 & 45.4 & 33.4 & 47.8 & 39.3 \\
InsTag-Diversity  & 43.4 & 53.4 & 46.1 & 35.0 & 49.1 & 40.1\\
\midrule 
\multicolumn{3}{l}{\it{\textbf{Prompt-and-Chosen-Response-Based Filtering}}} \\
IFD & 42.2 & 51.3 & 45.9 & 35.0 & 48.0 & 37.1\\
ppl(Chosen Response) & 43.4 & 52.5 & 47.0 & 37.2 & 49.4 & 37.6 \\
\midrule
\multicolumn{3}{l}{\it{\textbf{Chosen-Rejected-Response Based Filtering}}} \\
LLM-as-Prompt-Judge Pointwise  & 45.2 & 55.6 & 48.0 & 37.1 & 51.6 & 40.9\\
\method{}& 45.6 & 56.7 & 48.8 & 36.6 & 51.6 & 41.4 \\
\bottomrule
\end{tabular}
  \label{tab:wildbench_wildchat}
\end{table*}

\begin{table*}[t!]
 \small
 \setlength{\tabcolsep}{11pt}
\caption{{\bf Results of our DPO models trained with HelpSteer2.} Full AlpacaEval2 \& Arena-Hard Results of our DPO models trained with HelpSteer2 Dataset.}\label{tab:alpaca_arena_helpsteer2}
  \centering
\begin{tabular}{lrrrrrrrc}
\toprule
& \multicolumn{1}{c}{} &\multicolumn{3}{c}{AlpacaEval2}  & \multicolumn{3}{c}{Arena-Hard}  \\
\cmidrule(lr){3-5} \cmidrule(lr){6-8}
 & \multicolumn{1}{c}{Prompts} & \multicolumn{1}{c}{LC Win} & \multicolumn{1}{c}{Win} & \multicolumn{1}{c}{Len} & \multicolumn{1}{c}{Score}   & \multicolumn{1}{c}{95\% CI} & \multicolumn{1}{c}{Len}  \\
\midrule
\it{\textbf{Baseline}} \\
{Llama 3.1-8B-Instruct}  & - & 20.9& 21.8 &  2184 & 21.3 & (-1.9, 2.2) &  861\\
HelpSteer2 DPO (no filtering) & 10161 & 25.2 & 23.1 & 1733 & 26.8 & (-2.0, 2.4) & 606\\
\midrule
\it{\textbf{Prompt-Based-Filtering}} \\
LLM-as-Prompt-Judge Pointwise & 5376  & 27.8  & 25.7 & 1947 & 29.5 & (-2.8, 2.3) & 627 \\
\midrule
\it{\textbf{Prompt-Response-Based-Filtering}} \\
\method{}  & 5081  & 34.6 & 32.8  & 1941 &  35.0  & (-1.8, 2.2) & 621 \\
\bottomrule
\end{tabular}
\end{table*}

\begin{table*}[t!]
 \small
 \caption{{\bf Results on our DPO models trained with HelpSteer2.} Full WildBench results  of our DPO models trained with HelpSteer2 Dataset.}\label{tab:wildbench_eval_helpsteer}
  \centering
\begin{tabular}{lrrrrrrc}
\toprule
&\thead{WB-Score} & \multicolumn{5}{c}{WB-Score: Task-specific}  \\
\cmidrule(lr){2-2} \cmidrule(lr){3-7}
 & \thead{Weighted \\ average} & \multicolumn{1}{c}{Creative} & \thead{Planning  \& \\ Reasoning} & \thead{Math \& \\ Data Analysis}   & \thead{Information} & \thead{Coding \& \\ Debugging} 
\\
\midrule
\it{\textbf{Baseline}} \\
{Llama 3.1-8B-Instruct}  & 33.1 & 45.0 & 37.0 & 23.9 & 37.4 & 29.3\\
HelpSteer2 DPO (no filtering) & 37.1 & 48.6 & 40.4 & 26.5 & 44.3 & 33.4\\
\midrule
\it{\textbf{Prompt-Based-Filtering}} \\
LLM-as-Prompt-Judge Pointwise & 37.2 & 50.6 & 40.0 & 27.9 & 43.0 & 33.1\\
\midrule
\it{\textbf{Prompt-Response-Based-Filtering}} \\
\method{}  & 39.5 & 52.1 & 42.9 & 29.3 & 46.4 & 35.0\\
\bottomrule
\end{tabular}
\end{table*}

\begin{table*}[t!]
 \small
\setlength{\tabcolsep}{8pt}
\caption{{\bf Results of our DPO models trained with Self-Instructed Dataset:} Full AlpacaEval2 \& Arena-Hard Results comparing our method with training on standard Self-Instruct dataset.}
  \centering
\begin{tabular}{lrrrrrrrr}
\toprule
& & \multirow[b]{2}{4em}[-1.5mm]{\# Train examples} &\multicolumn{3}{c}{AlpacaEval2} & \multicolumn{3}{c}{Arena-Hard}  \\
\cmidrule(lr){4-6} \cmidrule(lr){7-9}
Training Prompts & \multicolumn{1}{c}{Filtering} &  & \multicolumn{1}{c}{LC Win} & \multicolumn{1}{c}{Win} & \multicolumn{1}{c}{Len} & \multicolumn{1}{c}{Score}   & \multicolumn{1}{c}{95\% CI} & \multicolumn{1}{c}{Len} \\
\midrule
Self-Instruct & None  & 20000 & 49.1 & 46.9 & 1956 &  38.5  & (-1.4, 1.6)  & 738 \\
Self-\method{} (without post-filtering) & None & 20000 & 53.6 & 56.1 & 2252 &  43.7 & (-2.3, 2.3)  & 777 \\
Self-Instruct with \method{} post-filtering & \method{} & 16261 & 58.3 & 53.2 & 1823 &   40.9& (-1.9, 1.6) & 560  \\
Self-\method{}  & \method{} & 18812 & {\bf 60.2} & {\bf 61.1} & 2121 &   42.1  & (-2.0, 2.4) & 606 \\
\bottomrule
\end{tabular}
  \label{tab:alpaca_arena_selfinstruct}
\end{table*}

\begin{table*}[t!]
 \small
 \caption{{\bf Results of our DPO models trained with Self-Instructed Dataset:} Full WildBench Results comparing our method with training on standard Self-Instruct dataset.}
\setlength{\tabcolsep}{4pt}
  \centering
\begin{tabular}{lrrrrrrrrc}
\toprule
& & \thead{WB-Score} & \multicolumn{5}{c}{WB-Score: Task-specific}  \\
\cmidrule(lr){3-3} \cmidrule(lr){4-8}
Training Prompts & \multicolumn{1}{c}{Filtering}  & \thead{Weighted \\ average} & \multicolumn{1}{c}{Creative} & \thead{Planning  \& \\ Reasoning} & \thead{Math \& \\ Data Analysis}   & \thead{Information} & \thead{Coding \& \\ Debugging} 
\\
\midrule
Self-Instruct & None  & 41.0 & 51.6 & 43.3 & 31.4 & 47.7 & 38.0\\
Self-\method{} (without post-filtering) & None & 44.8 & 55.3 & 46.9 & 33.5 & 49.7 & 44.6\\
Self-Instruct with \method{} post-filtering & \method{} & 44.1 & 54.8 & 47.3 & 36.4 & 48.2 & 40.3 \\
Self-\method{}  & \method{} & 42.5 & 54.1 & 46.2 & 32.8 & 48.6 & 38.2 \\
\bottomrule
\end{tabular}
  \label{tab:wildbench_eval_self_instr}
\end{table*}

\begin{table*}[t!]
 \small
 \caption{{\bf  Results of pair selections:} We report Armo scores on valid sets by varying different pairing methods instead of filtering prompts. Best pairing result 0.1842 is achieved with appointing response with bottom 25\% score as rejected, although still underperforming compared to our filtering method (0.1898).}
\setlength{\tabcolsep}{11pt}
  \centering
\begin{tabular}{lrr}
\toprule
Pair & Armo Score on Valid  \\
\midrule
Chosen=HighestScore, Rejected=LowestScore  & 0.1830 \\
Chosen=HighestScore, Rejected=BottomScore25\%   &  0.1842   \\
Chosen=HighestScore, Rejected=BottomScore50\%   & 0.1839      \\
Chosen=HighestScore, Rejected=BottomScore75\%   & 0.1821 \\
Chosen=HighestScore, Rejected=Random &  0.1835  \\
\midrule
Chosen=HighestScore, Rejected=LowestScore + \method{} & 0.1898 \\
\bottomrule   \\
\end{tabular}
  \label{tab:repair_ablation}
\end{table*}

\begin{table*}[t!]
 \small
 \caption{{\bf Results on Varying $N=8,16,32,64$ number of responses sampled per prompts in Response Generation}: Armo Score on Valid set of our DPO models trained with WildChat Dataset all increases after filtering based on \method{} regardless of the choice of $N$ in response generation step.}
\setlength{\tabcolsep}{11pt}
  \centering
\begin{tabular}{lrrc}
\toprule
 & \multicolumn{1}{c}{Before Filtering} &\multicolumn{1}{c}{After Filtering} & Gain \\
\emph{N} & \multicolumn{1}{c}{Armo Score} &  \multicolumn{1}{c}{Armo Score} \\
\midrule
8   & 0.1821 & 0.1860   & 0.0039 \\
16  &  0.1827 &  0.1878 & 0.0051  \\
32  &  0.1829 &  0.1882 & 0.0053 \\
64  & 0.1831  & 0.1898 & 0.0067 \\
\bottomrule
\end{tabular}
  \label{tab:subset_abalation}
\end{table*}

\subsection{t-SNE Analysis of Filtered Instructions}
\begin{figure*}[t!]
    \caption{
    {\bf t-SNE plots on instructions before and after filtering by rewards and lengths of rejected responses.} Red dots represent unfiltered instructions, while blue dots are instructions curated by filtering out those with low-reward and shorter rejected responses.
    }
    \includegraphics[width=0.45\linewidth]{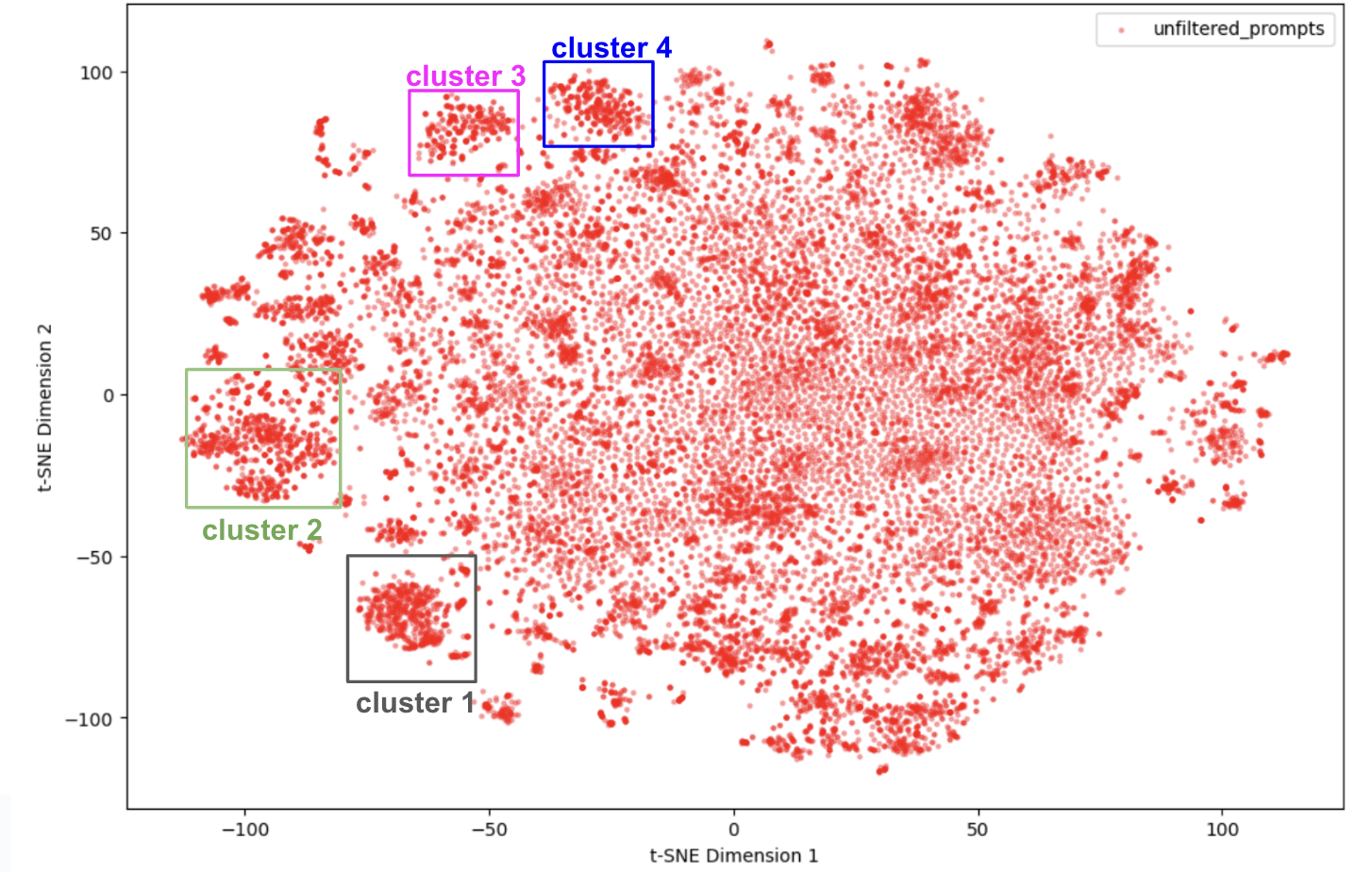}
    \includegraphics[width=0.45\linewidth]{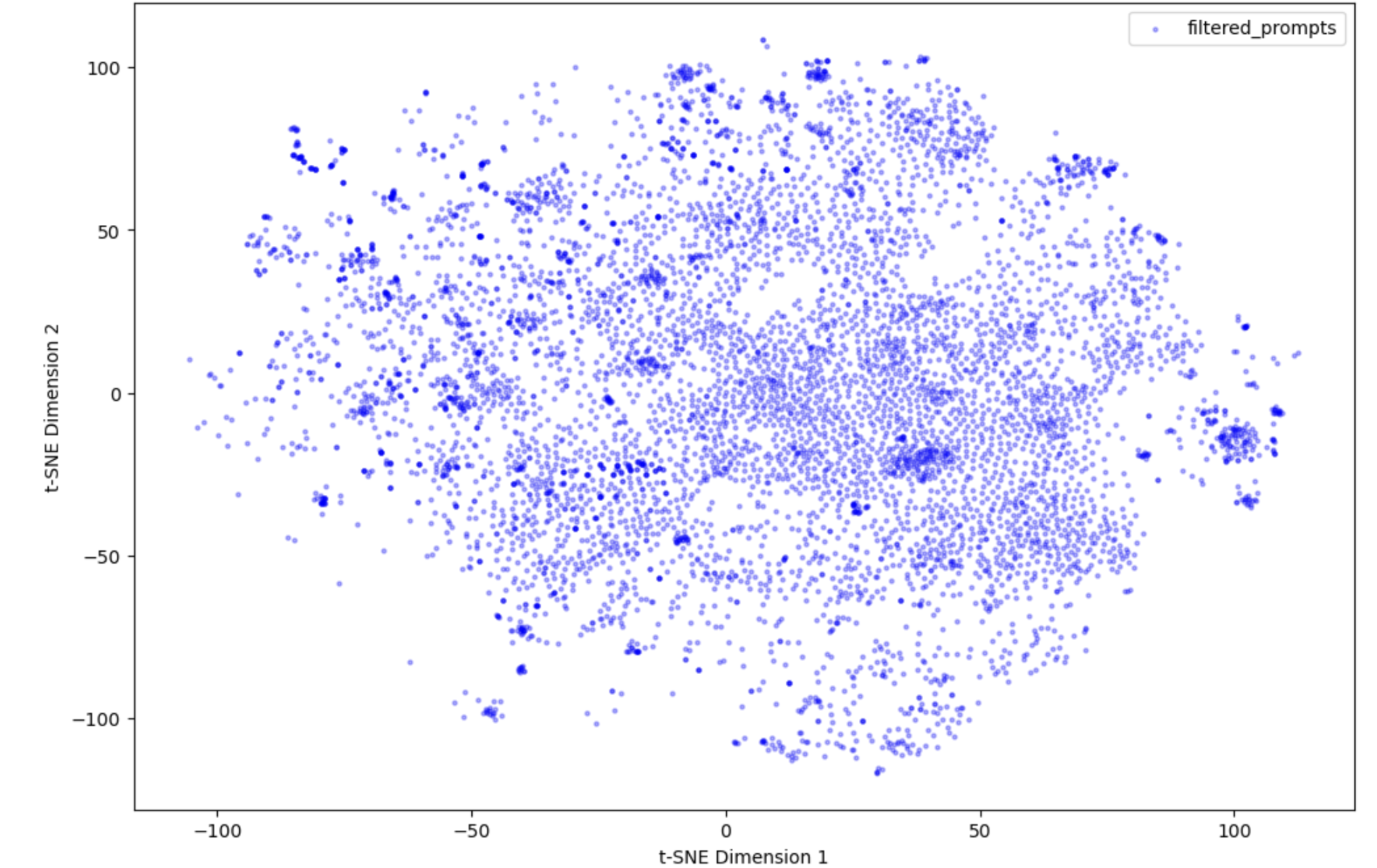}
    \hfill
    \label{fig:wildchat_unfiltere_vs_7k}
\end{figure*}

\begin{figure*}[t!]
    \caption{
    {\bf t-SNE plots on instructions before and after filtering by reward gaps.} Blue dots represent  instructions filtered only by rejected response, while yellow dots are instructions curated with smaller gap.
    }
    \includegraphics[width=0.48\linewidth]{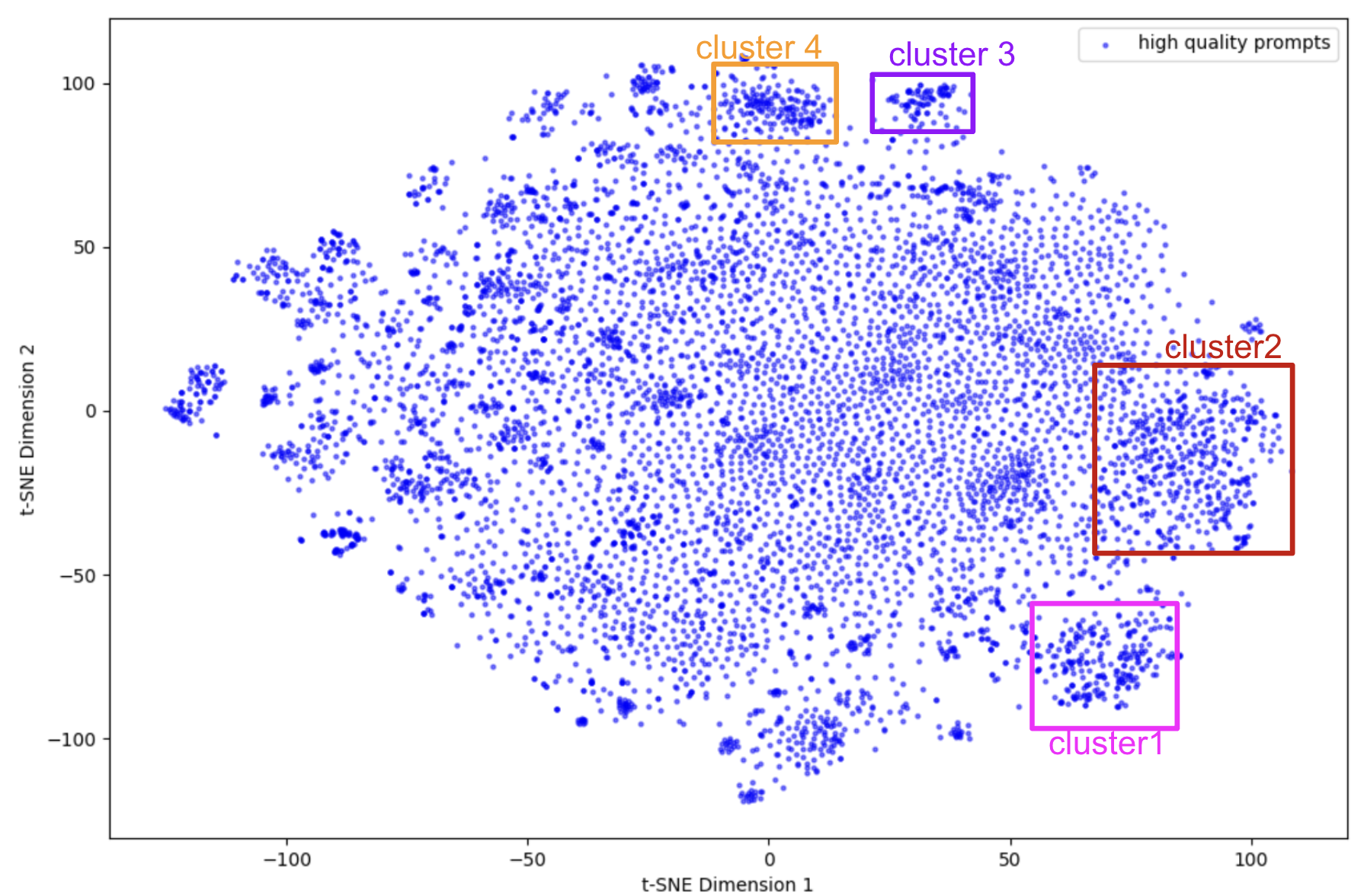}
    \includegraphics[width=0.48\linewidth]{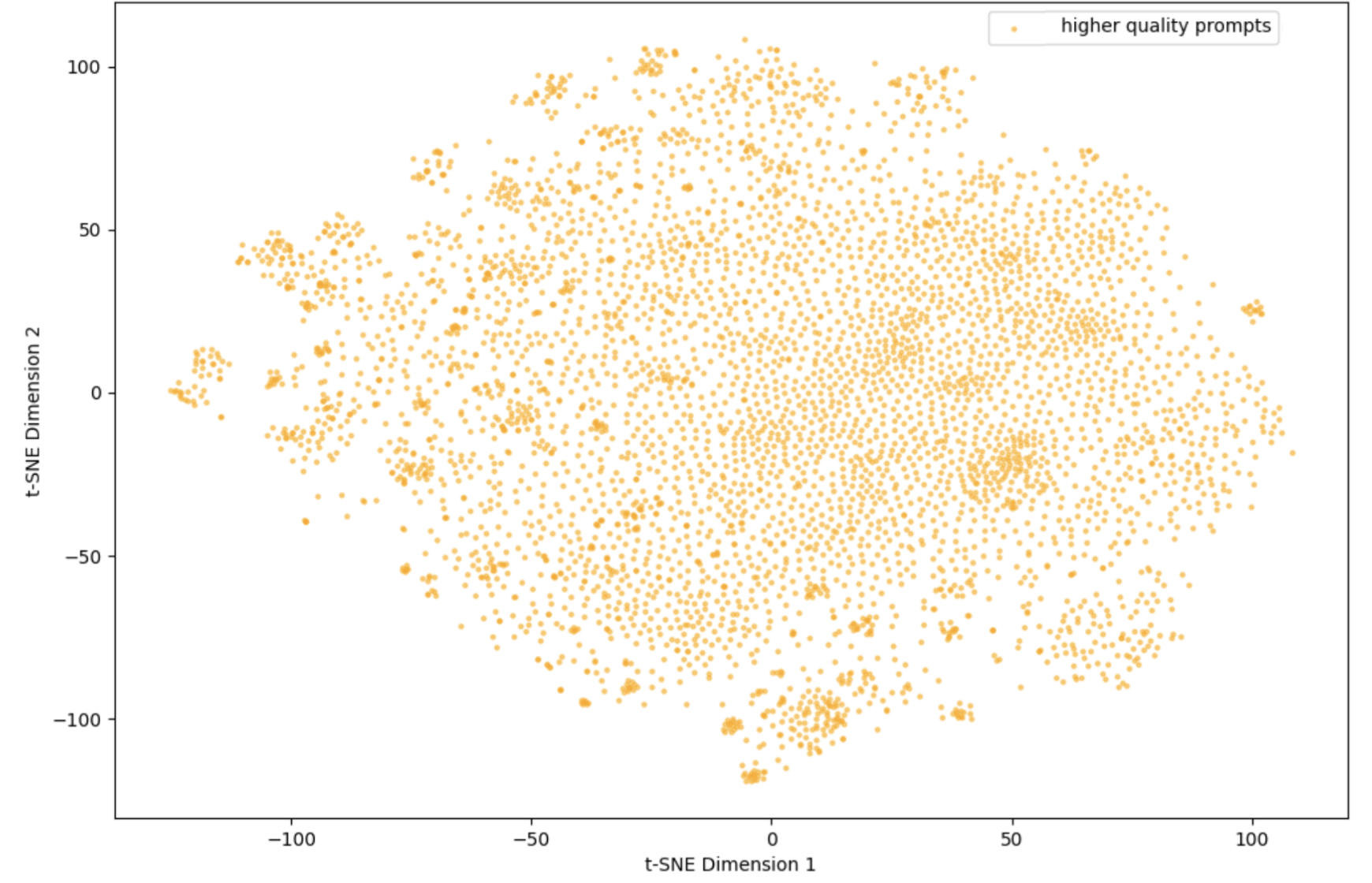}
    \hfill
    \label{fig:wildchat_7k_vs_5k}
\end{figure*}

We conduct t-SNE analysis on WildChat prompts filtered by rejected response length and reward in \autoref{fig:wildchat_unfiltere_vs_7k} and those further filtered by reward gap in \autoref{fig:wildchat_7k_vs_5k}. To better understand which prompts are being filtered out, we summarize prompts being filtered out due to rejected responses being of shorter length or lower reward
in \autoref{tab:wildchat_unfiltere_vs_7k_by_cluster}, and those filtered out due to large reward gaps in \autoref{tab:wildchat_7k_vs_5k_by_cluster}.

\subsection{Further Ablations}
\label{sec:ablation}
We report results of further ablation studies: comparing filtering and various pairing instead of filtering methods in \autoref{tab:repair_ablation}, and robustness of \method{} to choice of responses in rejection sampling in \autoref{tab:subset_abalation}.

\paragraph{\method{} outperforms alternative preference pairing methods}
We compare \method{} to methods without filtering that use different response pairing methods for building pairwise preferences. Recall that in our main experiments for \method{} we used the best-vs-worst pairing method as described in \autoref{sec:problem}.
Here we explore two alternative methods: (i) best-vs-random which is shown by existing work \citep{yasunaga2024alma,khaki2024rs} to outperform best-vs-worst,  and (ii) best-vs-bottom-K\% percentile where the rejected response has the bottom $K=25,50,75$ percentile score ($K=0$ being the lowest score). Both pairing methods can effectively lower reward gap and increase quality of rejected response without removing training prompts. We report model performance on the valid set in \autoref{tab:repair_ablation}. Out of all pairing methods, best-vs-bottom-25\%  works the best, but still under-performs compared with our \method{} method (pairing with best-vs-worst). When evaluated on AlpacaEval2, Arena-Hard, and WildBench, the model WildChat-20k DPO (best-vs-bottom-25\%) only achieves a slight improvement gain comparing to baseline WildChat-20k DPO (best-vs-worst), while still underperforming  \method{} as shown in \autoref{tab:alpaca_arena_wildbench_baseline}.
This result indicates that reward gap being small or the rejected reward being high  better works as an indication of a low-quality prompt rather than bad response pairing.

\paragraph{Combining alternative pairing with \method{} performs on par with best-vs-bottom pairing with \method{}.} We further apply \method{} filtering to examples paired by best-vs-bottom-25\% pairing. Combing best-vs-bottom-25\% with filtering out examples of low quality rejected responses yields ArmoRM Score of 0.18675, slightly lower than best-vs-worst + filtering by Rejected Reward (0.18795). Filtering out best-vs-bottom-25\% examples of bigger reward gaps yields to Armo Score of 0.1860 on valid set as compared to 0.18542 from best-vs-worst pairing + filtering by Reward Gaps.  Given the marginal performance gain between best-vs-worst and best-vs-bottom-25\% pairing with and without \method{}, we thus focus on the more widely adopted best-vs-worst pairing to experiment various filtering methods including our \method{} method.

\paragraph{\method{} is robust to the choice of the number of responses $N$.}
While we showed \method{} provides strong performance on HelpSteer2 where only $N=2$ responses are availabe for each prompt,  and on WildChat with $N=64$ responses sampled per prompt, 
we also compare the performance of \method{} by varying the choice of $N$ the number of candidate responses generated for preference annotations %
in the WildChat setup. As shown in \autoref{tab:subset_abalation}, for a wide range of values $N=64,32,16,8$, \method{} consistently outperforms the no filtering baseline, with larger $N$ achieving increasingly better performance, likely due to the increased quality and variability of chosen and rejected responses, allowing our \method{} metrics to be more accurate in curating high quality data.

\paragraph{Self-\method{} works with much smaller set of high-quality seed instructions} Instead of using all 4538 \method{} curated high-quality instructions as seed instructions $S$ during Step 1. few-shot generations, we sample a much shorter subset of 256 prompts from 4538 \method{}-curated prompts as seed instructions, and only conduct few-shot generations by sampling 8 prompts from the 256 seed prompts each time. We report Self-\method{} with and without post-filtering in \autoref{tab:alpaca_arena_wildbench_selfinstruct_fewshots}.  Self-\method{} based on 256 high-quality seed instructions (58.9) slightly underperforms than that based on 4538 seed prompts (60.2), but still outperforms Self-Instruct with \method{} post-filtering (58.3) as well as Self-\method{} based on all 4538 seed prompts without post-filtering (53.6), indicating that our method Self-\method{} can work well with a much smaller set of high-quality seed prompts.

\begin{table*}[t!]
 \small
 \caption{{\bf Self-\method{} for generating high-quality synthetic instructions by varying number of fewshots}. 
Self-\method{} creates prompts using few-shot samples from high-quality prompts curated by \method{}, whereas Self-Instruct uses few-shots from unfiltered WildChat prompts. Applying \method{} filtering {\em after} generation is also important, and achieves the best  results, significantly outperforming  Self-Instruct data.}
\setlength{\tabcolsep}{11pt}
  \centering
\begin{tabular}{lrccc}
\toprule
& \multirow[b]{2}{4em}[-1.5mm]{\# Seed \\ Prompts} & \multirow[b]{2}{4em}[-1.5mm]{\# Train examples}&\multicolumn{2}{c}{AlpacaEval2} \\
\cmidrule(lr){4-5}
{Train Prompts} &  & & \multicolumn{1}{c}{LC Win} & \multicolumn{1}{c}{Win} \\
\midrule
Llama 3.1-8B-Instruct (seed model) & - & - & 20.9 & 21.8  \\
 WildChat-20k + \method{} & - & 4538 &  57.8 & 57.2  \\
 \midrule
Self-Instruct + \method{} & 20000 & 16261 & 58.3 &3.2 \\
\midrule
Self-\method{} (without post-filtering) & 256  & 20000 & 50.0 & 51,2 \\
Self-\method{} (without post-filtering) & 4538    & 20000 & 53.6 & 56.1 \\
\midrule
Self-\method{} & 256   & 15619 & 58.9 & \textbf{63.1} \\
Self-\method{} & 4538  & 18812 & \textbf{60.2} & 61.1 \\
\bottomrule
\end{tabular}
  \label{tab:alpaca_arena_wildbench_selfinstruct_fewshots}
\end{table*}

\begin{table*}[ht]
    \caption{Performance of Different Filter Methods Across Quantile Ranges on WildChat with ArmoRM as reward anotator.}
    \centering
    \begin{tabular}{|l|c|c|c|c|c|c|}
        \hline
        Method & 0-100 & 10-100 & 25-100 & 50-100 & 60-100 & 75-100 \\
        \hline
        Chosen Reward & 0.18305 & \cellcolor{green!4}0.18325 & \cellcolor{green!21}0.18409 & \cellcolor{green!17}0.18393 & \cellcolor{green!15}0.18380 & \cellcolor{green!5}0.18333 \\
        \hline
        Rejected Reward & 0.18305 & \cellcolor{green!22}0.18411 & \cellcolor{green!21}0.18405 & \cellcolor{green!54}0.18566 & \cellcolor{green!100}0.18797 & \cellcolor{green!99}0.18795\\
        \hline
        Average Reward & 0.18305 & \cellcolor{green!12}0.18368 & \cellcolor{green!17}0.18392 & \cellcolor{green!38}0.18494 & \cellcolor{green!33}0.18468 & \cellcolor{green!27}0.18442\\
        \hline
        Chosen Length & 0.18305 & \cellcolor{green!9}0.18350 & \cellcolor{green!12}0.18366 & \cellcolor{red!8}0.18278 & \cellcolor{red!25}0.18226 & \cellcolor{red!64}0.18105\\
        \hline
        Rejected Length & 0.18305 & \cellcolor{green!14}0.18377 & \cellcolor{green!7}0.18340 & \cellcolor{green!54}0.18571 & \cellcolor{green!58}0.18593 & \cellcolor{green!34}0.18473\\
        \hline
    \end{tabular}
    \vspace{1em}
    
    \begin{tabular}{|c|c|c|c|c|}
        \hline
        Method & 0-100 & 0-25 & 0-50 & 50-100 \\
        \hline
        Reward Gap & 0.18305 & \cellcolor{green!20}0.18405 & \cellcolor{green!48}0.18542 & \cellcolor{red!100}0.17993 \\
        \hline
    \end{tabular}
    
    \label{table:wildchat_coordinate_wise}
\end{table*}

\begin{table*}[ht]
    \caption{Performance of Different Filter Methods Across Quantile Ranges on HelpSteer2 valid set.}
    \centering
    \begin{tabular}{|l|c|c|c|c|c|c|c|}
        \hline
        Method  & 0-25 & 0-50 & 0-75& 0-100 & 25-100 & 50-100 & 75-100 \\
        \hline
        Chosen Human Reward & \cellcolor{green!15}0.1469 & \cellcolor{green!4} 0.1451 & \cellcolor{green!4}0.1456  & 0.1458 & \cellcolor{green!12} 0.1461 & \cellcolor{green!4}0.1454 & \cellcolor{green!15} 0.1465 \\
        \hline
        Rejected Human Reward & \cellcolor{red!10} 0.1442 & \cellcolor{green!0}0.1455 & \cellcolor{green!0}0.1459  & 0.1458 & \cellcolor{green!25}0.1480 & \cellcolor{green!15}0.1470 & \cellcolor{green!15}0.1461 \\
        \hline        
        Chosen Length  & \cellcolor{green!30}0.1484 & \cellcolor{green!15}0.1467 & \cellcolor{green!0}0.1455  & 0.1458 & \cellcolor{green!0}0.1454 & \cellcolor{red!5}0.1446 & \cellcolor{red!5}0.1449 \\
        \hline
        Rejected Length   & \cellcolor{red!20}0.1421 & \cellcolor{red!15}0.1430 & \cellcolor{red!10}0.1445  & 0.1458 & \cellcolor{green!40}0.1495 & \cellcolor{green!50}0.1513 & \cellcolor{green!20}0.1478 \\
        \hline        
        Human Reward Gap  & \cellcolor{green!30}0.1482 & \cellcolor{green!30}0.1480 & \cellcolor{green!15}0.1466  & 0.1458 & \cellcolor{red!4}0.1448 & \cellcolor{red!5}0.1448 & \cellcolor{red!10}0.1441\\
        \hline
    \end{tabular}
    \vspace{1em}
\label{table:helpsteer_coordinate_wise}
\end{table*}

\begin{figure}[t]
\small
\caption{{\bf GPT4 eval prompt.}}
\centering
\begin{tcolorbox}[colback=green3!15!white, %
                  colframe=green3!40!white, %
                  arc=4mm, %
                  auto outer arc,
                  ]
I have a collection of prompts that I need to evaluate for their effectiveness in fine-tuning a language model.\\
A useful prompt should:\\
- Clearly ask a question\\
- Be concise and specific\\
- Directly relate to the topic of interest or follow given instructions\\
\\
Please assess each prompt and assign a score from 1 to 5 based on its usefulness:\\
- 1: Pretty useful\\
- 2: Somewhat useful\\
- 3: Neutral (neither useful nor harmful)\\
- 4: Somewhat harmful\\
- 5: Harmful\\
Make sure to clearly indicate the score at the end of your evaluation using the format: Score: x\\
Prompt: {\fontfamily{cmtt}\selectfont \bf\{prompt\}}\\
\end{tcolorbox}
\label{fig:gpt4_eval_prompt}
\end{figure}

\begin{figure}[t]
\small
\caption{{\bf Self-Instruct few-shot prompt template.}}
\centering
\begin{tcolorbox}[colback=green3!15!white, %
                  colframe=green3!40!white, %
                  arc=4mm, %
                  auto outer arc,
                  ]
Below are sample tasks from user.\\
1. \textless begin\textgreater{\fontfamily{cmtt}\selectfont \bf\{INSTRUCTION 1\}}\textless /end\textgreater\\\
2. \textless begin\textgreater{\fontfamily{cmtt}\selectfont \bf\{INSTRUCTION 2\}}\textless /end\textgreater\\
3. \textless begin\textgreater{\fontfamily{cmtt}\selectfont \bf\{INSTRUCTION 3\}}\textless /end\textgreater\\
4. \textless begin\textgreater{\fontfamily{cmtt}\selectfont \bf\{INSTRUCTION 4\}}\textless /end\textgreater\\
5. \textless begin\textgreater{\fontfamily{cmtt}\selectfont \bf\{INSTRUCTION 5\}}\textless /end\textgreater\\
6. \textless begin\textgreater{\fontfamily{cmtt}\selectfont \bf\{INSTRUCTION 6\}}\textless /end\textgreater\\
7. \textless begin\textgreater{\fontfamily{cmtt}\selectfont \bf\{INSTRUCTION 7\}}\textless /end\textgreater\\
8. \textless begin\textgreater{\fontfamily{cmtt}\selectfont \bf\{INSTRUCTION 8\}}\textless /end\textgreater\\
\\
Come up with a series of new tasks, wrapped with \textless begin\textgreater and \textless /end\textgreater\\
9. 
\end{tcolorbox}
\label{fig:fewshot_prompt}
\end{figure}

\subsection{Details about Baselines}
\label{app:baseline_details}

\paragraph{InsTag Complexity}

\citet{lu2023instag} utilized ChatGPT to generate semantic and intent-based tags, which were then used to fine-tune a large language model (LLM) data\
 tagger. The number of tags per prompt served as a complexity metric. Building on their methodology, we employed a publicly available tagger
(\url{https://github.com/OFA-Sys/InsTag}. Note that Meta was not involved in the training of the Instag model we used.)
to annotate each prompt, generating between 1 and 100 tags per prompt. We then categorized our training prompts into four groups based on the number of tags: more than 2, more than 3, more than 4, and more than 5. From each group, we randomly sampled 10,000 training data samples and trained a distinct model for each group. It is important to note that a threshold of $\ge 1$ implies no filtering, with only a random sample of 10,000 data points from WildChat. A threshold of $\ge 2$ means filtering out prompts with only 1 tag. As shown in \autoref{tab:instag_complexity}, a threshold of $\ge 2$ yields the best performance using the InsTag Complexity filtering method. We reported the results for a threshold of $\ge 2$ in \autoref{tab:alpaca_arena_wildbench_baseline}.

\begin{table}[ht]
  \centering
  \begin{minipage}{0.45\textwidth}
    \centering
    \small
        \caption{Model performance with InsTag Complexity Filtering}
    \begin{tabular}{c|>{\centering\arraybackslash}p{18mm}}
    \toprule
        Tag threshold & Armo Reward on Valid Set \\
        \hline
         $>= 1$ & 0.1820 \\
         $>= 2$ & 0.1826 \\
         $>= 3$ & 0.1812 \\
         $>= 4$ & 0.1815 \\ 
         $>= 5$ & 0.1818 \\
         \bottomrule
    \end{tabular}
    \label{tab:instag_complexity}
  \end{minipage}%
  \hspace{0.03\textwidth} 
  \begin{minipage}{0.45\textwidth}
    \centering
    \small
        \caption{Model performance with InsTag Diversity Filtering}
    \begin{tabular}{>{\centering\arraybackslash}p{12mm}|>{\centering\arraybackslash}p{12mm}|>{\centering\arraybackslash}p{18mm}}
    \toprule
        Tag Frequency & Max prompt per Tag &Armo Reward on Valid Set \\
        \hline
         1 & 1& 0.1809 \\
         1 & 2 & 0.1799 \\
         2 & 1 & 0.1798 \\
         2 & 2 & 0.1800 \\
         3 & 1 & 0.1799 \\
         3 & 2 & 0.1797 \\
         3 & 3 & 0.1814 \\
         4 & 3 & 0.1821 \\
         5 & 3 & 0.1818 \\
         6 & 3 & 0.1831 \\
         \bottomrule
    \end{tabular}
    \label{tab:instag_diversity}
  \end{minipage}
\end{table}

\paragraph{InsTag Diversity}

The InsTag Diversity filtering method \citep{lu2023instag} considers a dataset to be more diverse if it contains a larger number of unique tags, as annotated by the aforementioned tagger. We employed two metrics to manage InsTag Diversity:

1. Tag Frequency: We deem a tag valid if it meets a predefined frequency threshold. This approach addresses the issue of infrequent tags, such as ``serve size'' and ``market failure,'' which appeared only once or twice in the entire Wildchat dataset, suggesting they may not represent valid categories. In contrast, more common tags like ``creative writing'' and ``information retrieval'' are more appropriate for categorizing prompt data.

2. Max prompt per Tag: This metric controls the coverage ratio of unique tags. If a prompt contains only tags that have already been covered by the selected set, we discard the prompt to ensure diversity.

\autoref{tab:instag_diversity} presents the performance results when diversity is controlled using the two metrics described above. To ensure fairness, we downsampled the training data for each experiment to 10,000 samples. The results indicate that the model achieves optimal performance when the Tag Frequency is set to 6 and the Max Prompt per Tag is set to 3. This means we only consider tags that appear more than six times in the entire Wildchat dataset, and we allow a maximum of three prompts per tag. The best performance results are reported in \autoref{tab:alpaca_arena_wildbench_baseline}.

\paragraph{Perplexity}

To curate training prompts, we compute the perplexity (ppl) of the selected response $y_w$ using the Llama-3.1-8B-Instruct model in a zero-shot setting. We use this perplexity as a filtering metric, specifically retaining examples with high $\text{ppl}(y_w|x)$ values, which may indicate more challenging prompts. We adjust the quantile range to control perplexity, calculating $\text{ppl}(y_w|x)$ for 20,000 Wildchat data points and filtering them based on this range. \autoref{tab:ppl} displays model performance across different ppl quantile ranges. As shown, the quantile range of 25-100 yields the best performance, and we report this model's performance in \autoref{tab:alpaca_arena_wildbench_baseline}.

\begin{table}[t!]
  \centering
  \begin{minipage}{0.45\textwidth}
    \centering
    \small
        \caption{Model performance with Perplexity Filtering}
    \begin{tabular}{c|c}
    \toprule
        Quantile Range & Armo Reward on Valid Set \\
        \hline
         25-100 & 0.1833 \\
         50-100 & 0.1827 \\
         75-100 & 0.1797 \\
         \bottomrule
    \end{tabular}
    \label{tab:ppl}
  \end{minipage}%
  \hspace{0.03\textwidth} 
  \begin{minipage}{0.45\textwidth}
    \centering
    \small
        \caption{Model performance with IFD Filtering}
    \begin{tabular}{c|c}
    \toprule
        Quantile Range & Armo Reward on Valid Set \\
        \hline
         0-25 & 0.1815 \\
         0-50 & 0.1823 \\
         0-75 & 0.1832\\
         25-100 & 0.1835 \\
         \bottomrule
    \end{tabular}
    \label{tab:IFD}
  \end{minipage}
\end{table}

\paragraph{Instruction-Following Difficulty (IFD)}

\citet{li2023quantity} introduced the IFD to measure the model-specific difficulty of a data sample. In the instruction-tuning process, the loss of a sample pair (Q, A) is calculated by continuously predicting the next tokens given the instruction Q and their proceeding words:

\begin{equation}
L_\theta(A \mid Q)=-\frac{1}{N} \sum_{i=1}^N \log P\left(w_i^A \mid Q, w_1^A, w_2^A, \ldots, w_{i-1}^A ; \theta\right)
\end{equation}

where N is the number of words of the groundtruth answer A. They denote this averaged crossentropy loss as the Conditioned Answer Score
$S_\theta(A \mid Q) = L_\theta(A \mid Q)$.

Then they introduce the Direct Answer Score $S_\theta(A)$

\begin{equation}
s_\theta(A)=-\frac{1}{N} \sum_{i=1}^N \log P\left(w_i^A \mid w_1^A, \ldots, w_{i-1}^A ; \theta\right)
\end{equation}

Finally, they estimate the Instruction-Following Difficulty (IFD) scores $IFD_\theta(Q,A)$ on following instruction of a given (Q, A) pairs by calculating the ratio between $S_\theta(A)$
and $S_\theta(A\mid Q)$:

\begin{equation}
\operatorname{IFD}_\theta(Q, A)=\frac{s_\theta(A \mid Q)}{s_\theta(A)}
\end{equation}

We calculated the IFD scores for 20,000 Wildchat data points and filtered them based on specific ranges. As shown in \autoref{tab:IFD}, filtering with a range of 25-100 yielded the best performance. The performance of this model is reported in \autoref{tab:alpaca_arena_wildbench_baseline}.

\begin{table*}[t!]
 \small
 \caption{{\bf Noisy instruction filtered based on rejected responses of lower scores and shorter lengths.} We expand 4 clusters of instructions highlighted in \autoref{fig:wildchat_unfiltere_vs_7k} for a better understanding of what instructions are being filtered out by measuring quality of rejected responses.}\label{tab:wildchat_unfiltere_vs_7k_by_cluster}
\setlength{\tabcolsep}{11pt}
  \centering
\begin{tabular}{p{1cm}|p{4cm}|p{5cm}|p{4cm}}
\toprule
Cluster & Description &  Rejecting Reason & Rejected Instruction \\
\midrule
Cluster 1 & 646 instructions in the format of {\it``give me a response to ```$<$text$>$``` to send in a discussion, VERY SHORT, CONCISE \& CLEAR. ONLY RETURN THE RAW MESSAGE, DO NOT SAY "Hey here is the message you asked"''}, where {\it $<$text$>$} refers to a single-turn conversational message. & Around 90\% of rejected responses are of shorter lengths and lower scores below 25\% percentile, even though their scores are higher than average rejected scores. These short and concise conversational responses are shorter thus potentially more generic and less informative for the models to further improve upon. & give me a response to ```I'm feeling great! Swimming around in the ocean and hunting for prey never gets old. I'm always looking for new and exciting ways to keep busy.``` to send in a discussion, VERY SHORT, CONCISE \& CLEAR. ONLY RETURN THE RAW MESSAGE, DO NOT SAY ``Hey here is the message you asked''\\
\midrule
Cluster 2 & 804 instructions in the format of movie script: {\it (In a $<$scene$>$) $<$name1$>:<$line1$>$\escape{n}... $<$nameK$>:<$lineK$>$}, without any instructions on what the model response should be. & {\bf Short rejected responses and low rejected scores:} Around 90\% of rejected responses are of shorter lengths, lower scores below 50\% percentile. In addition, over 75\% response pairs are of larger score gap above 50\% percentile. All of these are likely due to the obscurity of the user instructions. & (In the school literature clubroom...)\escape{n}\escape{n}Monika: Natsuki, where is everyone? I haven't seen Sayori, Yuri, or MC in a while.\escape{n}Natsuki:... \\ 
\midrule
Cluster 3 & 279 instructions, majority of them are purely excerpts from a fictional story, with no specifications on what the response should be. Users could be asking models to continue the story, or summarize, or edit it. & {\bf Short rejected responses and low rejected scores:} Over 90\% of rejected responses are of shorter lengths, lower scores below 50\% percentile. All of these are likely due to the obscurity of the user instructions. & David insists he is too strong-willed and intelligent to ever be hypnotized. He scoffs at the very idea. In this kinky script, his colleague Clare easily proves him wrong, in front of some amused co-workers.\\
\midrule
Cluster 4 & 466 instructions, majority of them are about writing a comedic story about a fictional character. & {\bf Short rejected responses and low rejected scores:} Around 95\% of rejected responses are of shorter lengths, lower scores below 25\% percentile. All of these are likely due to the Llama 3.1-8B-Instruct model being reluctant to provide detailed answers. These instructions are therefore less informative for improving Llama 3.1-8B-Instruct with its own responses. & Make a story about Shrek in the buff and farting in bog water, then collecting all the fish the smell kills and eating them for dinner. \\
\bottomrule
\end{tabular}
\end{table*}

\begin{table*}[t!]
 \small
 \caption{{\bf Noisy instruction clusters filtered based on rejected responses of lower scores and shorter lengths.} We expand 4 clusters of instructions sampled from \autoref{fig:wildchat_7k_vs_5k}, that consists of both rejected and accepted instructions by \method{}.}
\setlength{\tabcolsep}{11pt}
  \centering
\begin{tabular}{p{1cm}|p{2.5cm}|p{2.5cm}|p{3cm}|p{3.5cm}}
\toprule
Cluster & Description & Rejecting Reason & Rejected Instructions & Accepted  Instructions \\
\midrule
Cluster 1 & 140 instructions, among those some are purely code snippets without additional guidelines on what to respond. Others are requests asking to optimize a given piece of code. & Instructions with pure code snippets lead to variable responses (from code refactoring, editing, code completion, to code review and code explanation using natural language). Instructions on ``improve this code'' can also incur variable responses given the lack of more specified instructions. & improve this emergency shutdown code:  
import os\escape{n}import platform\escape{n}import sys\escape{n}import secrets\escape{n}from threading import Thread, Event\escape{n}from pynput.mouse import Listener\escape{n}from pynput.keyboard... & I will provide you disassembly from a computer game that runs in MS-DOS. The game was written in C with a Watcom compiler. Some library function calls are already identified. Explain the functions I give to you and suggest names and C-language function signatures for them, including which parameters map to which registers or stack values <code snippet>....\\
\midrule
Cluster 2 & 237 instructions including: writing a program, inquiry about online tool, software installation, etc. Many  instructions are short (in 120 characters) and relatively high-level.  &  Rejected responses are on average much longer and complex compared to chosen responses, despite the high scores of both chosen and rejected responses. & alignment in excel vb.net \newline \newline write script for delegating fb  group & i am getting access denined when i try to put local files into remote server using ftp how can i resolve this issue \\ 
\midrule
Cluster 3 \& 4 & Cluster 3 are 52 instructions related to hypothetical or surreal scenarios; Cluster 4 are 52 instructions in the form of ``Freedom planet ...'', possibly for a creative project in the video game. & Model responses vary a lot due to the obscurity or hypothetical nature of the instructions.  & Can You Imagine 4 Fictional Versions Of Silicon Valley During 1940 In Detail? \newline \newline Freedom planet and Madness combat all characters: Hank 4th wall breaks and repetition & What if Cartoon Network Made The Amazing World of Gumball: Next Generation \newline \newline freedom planet what if Lord Brevon Wins (not kils Lilac, Carol and Milla) \\
\bottomrule
\end{tabular}
  \label{tab:wildchat_7k_vs_5k_by_cluster}
\end{table*}

\end{document}